\documentclass[lettersize,journal]{IEEEtran}
\usepackage{amsmath,amsfonts}
\usepackage{algorithmic}
\usepackage{algorithm}
\usepackage{array}
\usepackage[caption=false,font=normalsize,labelfont=sf,textfont=sf]{subfig}
\usepackage{textcomp}
\usepackage{stfloats}
\usepackage{url}
\usepackage{verbatim}
\usepackage{graphicx}
\usepackage{cite}
\usepackage{graphicx}
\usepackage{amsmath}
\usepackage{amssymb}

\usepackage{xcolor}
\usepackage{enumitem}
\usepackage{xspace}
\usepackage{booktabs}
\usepackage{colortbl}
\usepackage{multirow}
\usepackage{makecell}
\usepackage{lipsum} %
\usepackage{gensymb} %
\usepackage[pagebackref,breaklinks,colorlinks,linkcolor={red!50!black},citecolor={green!50!black}]{hyperref}

\newcommand{\Tref}[1]{Table~\ref{#1}}
\newcommand{\eref}[1]{Eq.~\eqref{#1}}

\newcommand{\fref}[1]{Fig.~\ref{#1}}
\newcommand{\Fref}[1]{Figure~\ref{#1}}
\newcommand{\sref}[1]{Sec.~\ref{#1}}

\usepackage[labelsep=period]{caption}
\renewcommand{\paragraph}[1]{\vspace{0.2em}\noindent \textbf{#1 \hspace{0.2em}}}

\definecolor{MyDarkRed}{rgb}{0.56, 0.16, 0.16}
\definecolor{MyBrightRed}{rgb}{0.86, 0.16, 0.16}
\definecolor{MyDarkBlue}{rgb}{0.16, 0.16, 0.66}

\newcommand{\delayedGS}{delayed Gaussian growth\xspace}
\newcommand{\DelayedGS}{Delayed Gaussian Growth\xspace}
\newcommand{\coarsetofine}{scale-cascaded mask bootstrapping\xspace}
\newcommand{\CoarseToFine}{Scale-cascaded Mask Bootstrapping\xspace}

\newcommand{\model}{\mathcal{G}}
\newcommand{\element}{g}

\newcommand{\opacity}{\alpha}

\newcommand{\gcolor}{\mathbf{c}}
\newcommand{\loss}{\mathcal{L}}

\begin{document}

\title{RobustSplat++: Decoupling Densification, Dynamics, and Illumination for In-the-Wild 3DGS}

\author{
    Chuanyu Fu,
    Guanying Chen, %
    Yuqi Zhang,
    Kunbin Yao,
    Yuan Xiong,
    Chuan Huang,\\ %
    Shuguang Cui,~\IEEEmembership{Fellow,~IEEE,}
    Yasuyuki Matsushita,~\IEEEmembership{Fellow,~IEEE,}
    Xiaochun Cao%
}

\markboth{Journal of \LaTeX\ Class Files,~Vol.~14, No.~8, August~2021}%
{Shell \MakeLowercase{\textit{et al.}}: A Sample Article Using IEEEtran.cls for IEEE Journals}

\IEEEpubid{0000--0000/00\$00.00~\copyright~2021 IEEE}

\IEEEaftertitletext{
\begin{center}
    \includegraphics[width=\textwidth]{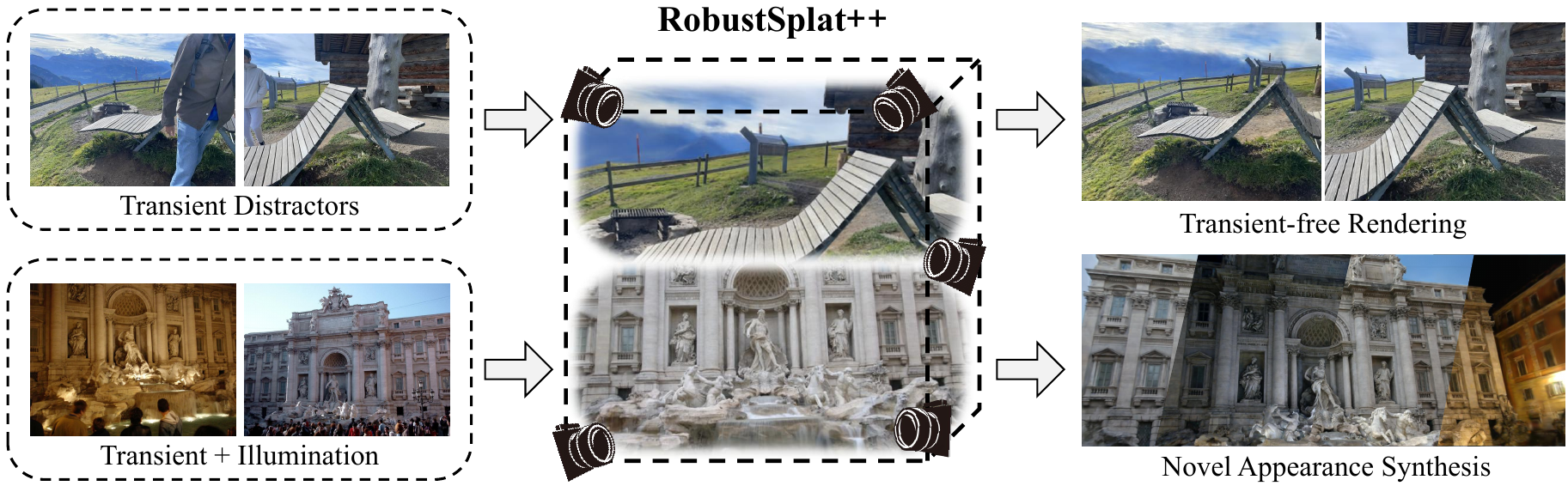}
    \captionof{figure}{We propose a robust solution, \emph{RobustSplat}{++}, to handle 3DGS optimization in in-the-wild scenes. {Our method can effectively handle transient distractors alone or in combination with illumination variations, yielding clean and more reliable results.}}
    \label{fig:teaser}
\end{center}
}
\maketitle

\begin{abstract}
    3D Gaussian Splatting (3DGS) has gained significant attention for its real-time, photo-realistic rendering in novel-view synthesis and 3D modeling. 
    However, existing methods struggle with accurately modeling {in-the-wild} scenes affected by transient objects and illuminations, leading to artifacts in the rendered images.
    We identify that the Gaussian densification process, while enhancing scene detail capture, unintentionally contributes to these artifacts by growing additional Gaussians that model transient disturbances {and illumination variations}.
    To address this, we propose RobustSplat++, a robust solution based on several critical designs.
    First, we introduce a delayed Gaussian growth strategy that prioritizes optimizing static scene structure before allowing Gaussian splitting/cloning, mitigating overfitting to transient objects in early optimization. 
    Second, we design a scale-cascaded mask bootstrapping approach that first leverages lower-resolution feature similarity supervision for reliable initial transient mask estimation, taking advantage of its stronger semantic consistency and robustness to noise, and then progresses to high-resolution supervision to achieve more precise mask prediction.
    Third, we incorporate the delayed Gaussian growth strategy and mask bootstrapping with appearance modeling to handling in-the-wild scenes including transients and illuminations.
    Extensive experiments on multiple challenging datasets show that our method outperforms existing methods, clearly demonstrating the robustness and effectiveness of our method. 
    Our code, datasets, and models are available at \href{https://fcyycf.github.io/RobustSplat-plusplus/}{https://fcyycf.github.io/RobustSplat-plusplus/}.
\end{abstract}


\begin{figure*}[ht] 
\begin{center}
    \input{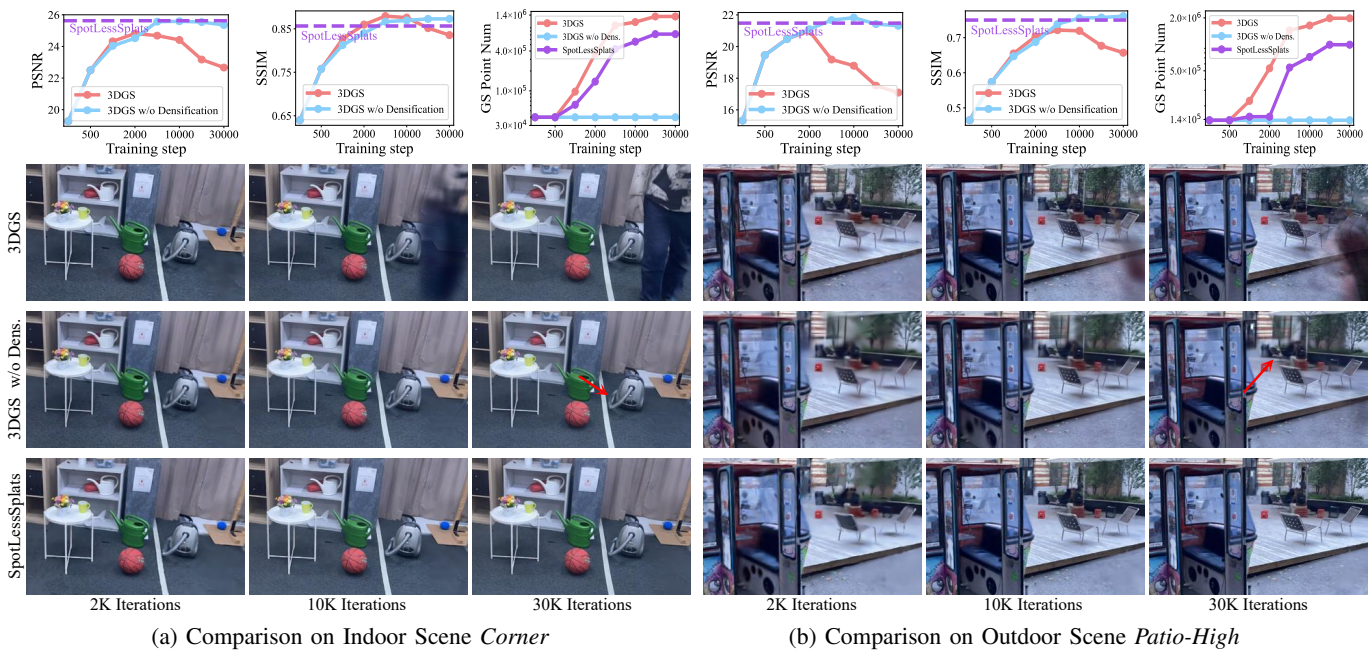}
    \caption{\textbf{Analysis of Gaussian densification in transient object fitting.} 
    As training progresses, vanilla 3DGS~\cite{kerbl20233d} suffers from performance degradation and exhibits artifacts due to the increasing number of Gaussians. Disabling Gaussian densification notably improves the results, even achieving performance comparable to the recent robust method SpotLessSplats~\cite{sabour2024spotlesssplats}. 
    Despite producing transient-free rendering, \emph{3DGS w/o densification} struggles to recover fine details in regions with sparse Gaussian initialization (highlighted by \textcolor{MyDarkRed}{red arrows}).}
    \label{fig:dens_analysis}
\end{center} 
\end{figure*}

\section{Introduction}
\label{sec:intro}

\IEEEPARstart{S}{ignificant} advancements have been made recently in novel-view synthesis and 3D reconstruction from multi-view images \cite{mildenhall2020nerf,tewari2021advances,wu2024recent}. 
Among these, 3D Gaussian Splatting (3DGS) stands out as an effective approach, enabling real-time and realistic rendering~\cite{kerbl20233d}. 
The optimization of 3DGS starts from a sparse set of points obtained through Structure-from-Motion (SfM), and adaptively controls the number and density of Gaussians to create an accurate 3D representation.
To capture fine details, the Gaussians will be split or cloned when the accumulated gradient magnitude of their centered position exceeds a predefined threshold.

However, existing methods often assume static scene conditions, an assumption frequently violated in real-world scenarios containing transient objects {and illumination variations}. 
These mismatches break the multi-view consistency requirement, leading to severe artifacts and degraded reconstruction quality~\cite{kulhanek2024wildgaussians}.

\IEEEpubidadjcol

\paragraph{Challenges of Transient-free 3DGS}
The key challenge lies in accurately detecting and filtering motion-affected regions across different images. Existing approaches primarily follow three paradigms:
(1) category-specific semantic masking (e.g., humans and vehicles), which struggles to generalize to diverse transient objects;
(2) uncertainty-based masking derived from photometric reconstruction loss minimization, but it often fails to reliably predict motion masks~\cite{martin2021nerf}; 
and (3) learning-based motion masking, where an MLP predicts motion masks using image features (e.g., DINO features~\cite{oquab2023dinov2}) as input and is supervised by photometric residuals~\cite{sabour2024spotlesssplats} or feature similarity~\cite{kulhanek2024wildgaussians,goli2024romo} between captured and rendered images.

While learning-based methods have shown strong performance in transient-free 3DGS optimization, they face critical limitations. 
In the early stages of training, the 3DGS representation is under-optimized, resulting in over-smooth renderings with large photometric residuals and weak feature similarity in both dynamic and static regions. 
Using these unreliable signals as supervision for mask estimation leads to inaccurate transient masks, with small masks failing to remove transients and causing artifacts, while overly smooth early reconstructions misclassify static regions, hindering optimization and resulting in under-reconstruction.

\paragraph{Analysis} 
To mitigate these issues, two critical aspects need to be considered. 
First, the optimization of 3DGS should be explicitly constrained from overfitting to transient regions without accurate transient masks during the initial optimization phases.
Second, mask supervision in the early iterations should be designed to tolerate under-reconstructed regions, allowing better reconstruction of static areas.

Through a detailed analysis of the 3DGS method, we identify that the Gaussian densification process (which, by default, begins after 500 iterations) enhances scene detail capture but unintentionally introduces artifacts (see~\fref{fig:dens_analysis}). 
Initially, 3DGS fits the static parts of the scene well; however, as densification progresses, it tends to overfit dynamic regions, resulting in artifacts in areas influenced by moving objects. 
Surprisingly, we find that \emph{explicitly disabling the densification process in vanilla 3DGS effectively mitigates these artifacts}, yielding results comparable to SpotLessSplats~\cite{sabour2024spotlesssplats} without requiring any specialized design. 

This is because, without densification, the image reconstruction loss provides limited positional gradients for 3D Gaussians, primarily optimizing their shape and color instead. As a result, the initially placed Gaussians remain stable in position, reducing the risk of overfitting to transient elements.  
However, the absence of densification leads to an insufficient number of Gaussians to represent fine details, causing the rendered images to appear overly smooth in regions with sparse point initializations.

Notably, we extend our analysis to in-the-wild scenes with lighting variations and observe consistent findings. The detailed analysis will be discussed in \sref{sec:method_illumination}.

\paragraph{The Proposed Solution}
Building on our analysis, we propose a simple yet effective method, called \emph{RobustSplat}, {to optimize 3DGS for handling transient challenges in in-the-wild scenes.} 
Our method introduces two key designs. 
First, we propose a \emph{\delayedGS} strategy that prioritizes reconstructing the global structure of the 3D scene while explicitly avoiding premature fitting to dynamic regions.
Second, to improve the mask supervision signal for under-reconstructed regions while preserving sensitivity to transient areas, we introduce a \emph{\coarsetofine} approach. This approach progressively increases the supervision resolution, leveraging the observations that low-resolution features capture global consistency more effectively and suppressing local noise in early optimization stages.

To handle the illumination challenges in in-the-wild scenes, we further investigate how illumination variations affect densification processing, and extend \emph{RobustSplat} by integrating it with appearance modeling, resulting in an enhanced method called \emph{RobustSplat}++.

In summary, our key contributions are:  
\begin{itemize}[itemsep=0pt,parsep=0pt,topsep=2bp]
    \item We analyze how the 3DGS densification process contributes to artifacts caused by transient objects, offering new insights for improving the optimization of distractor-free 3DGS.
    \item We propose \emph{RobustSplat}, a robust approach that integrates the \delayedGS strategy and \coarsetofine to effectively reduce the impact of dynamic objects during 3DGS optimization.
    \item We propose extended appearance modeling for RobustSplat, denoted as \emph{RobustSplat}++, to handle the cases where scenes exhibit illumination variations.
    \item We demonstrate that our approach outperforms state-of-the-art methods with a simple yet effective design.
\end{itemize}

We have presented preliminary results of this work in~\cite{2025robustsplat}, which is extended in this paper in several aspects. 
First, we present more detailed ablation studies and experimental results for transient-free 3DGS optimization.
Second, we provide a comprehensive analysis of illumination variation challenges and demonstrate their detrimental impact on 3DGS optimization stability.
Third, we incorporate the \delayedGS strategy with appearance modeling to address the limitation of RobustSplat for illumination variations, which enables extension to more generalized in-the-wild data. 
Lastly, we provide a comprehensive comparison between our method and the recent state-of-the-art methods.

\section{Related Work}
\label{sec:related_works}

\paragraph{Novel View Synthesis} 
Neural radiance field (NeRF)~\cite{mildenhall2020nerf}, as a representative approach for novel view synthesis, is widely recognized for its highly realistic rendering capabilities~\cite{tewari2021advances, tewari2020state, tang2022compressible, xie2022neural}. 
Many follow-up NeRF-based methods have introduced numerous enhancements in terms of efficiency~\cite{muller2022instant, chen2022tensorf, fridovich2022plenoxels} and performance~\cite{barron2022mip, lu2023large, mari2022sat, oechsle2021unisurf, wang2021neus, wu2022scalable, xu2023grid, yu2022monosdf, zhang2024aerial,yang2023freenerf}, achieving highly effective results. Recently, a novel explicit representation, 3D Gaussian Splatting (3DGS)~\cite{kerbl20233d}, has sparked considerable attention for its transformative impact on novel view synthesis methods due to its real-time rendering capability~\cite{yu2024mip, guedon2024sugar, lin2024vastgaussian, zhou2024hugs}.

\paragraph{Robustness in NeRF}
The vanilla NeRF assumes a static scene, but this assumption often fails with in-the-wild situations, where unconstrained images inevitably include lighting variations and dynamic/transient objects. 
NeRF-W~\cite{martin2021nerf} introduces an appearance embedding for exposure and transient modeling, which has been widely used~\cite{yang2023cross,chen2022hallucinated}. 
For distractor removal, it uses MLPs to predict uncertainty and following methods~\cite{lee2023semantic, ren2024nerf} introduce features from large pre-trained models~\cite{caron2021emerging, oquab2023dinov2} to improve robustness.
Another branch, represented by RobustNeRF~\cite{sabour2023robustnerf}, utilizes image residuals to predict binary masks for dynamic objects, filtering them out during training~\cite{otonari2024entity, chen2024nerf}.
Moreover, $D^{2}$-NeRF~\cite{wu2022d} decouples a dynamic scene into three fields, including static field, dynamic field, and non-static shadow field.

\begin{figure*}[tb] \centering
    \includegraphics[width=0.98\textwidth]{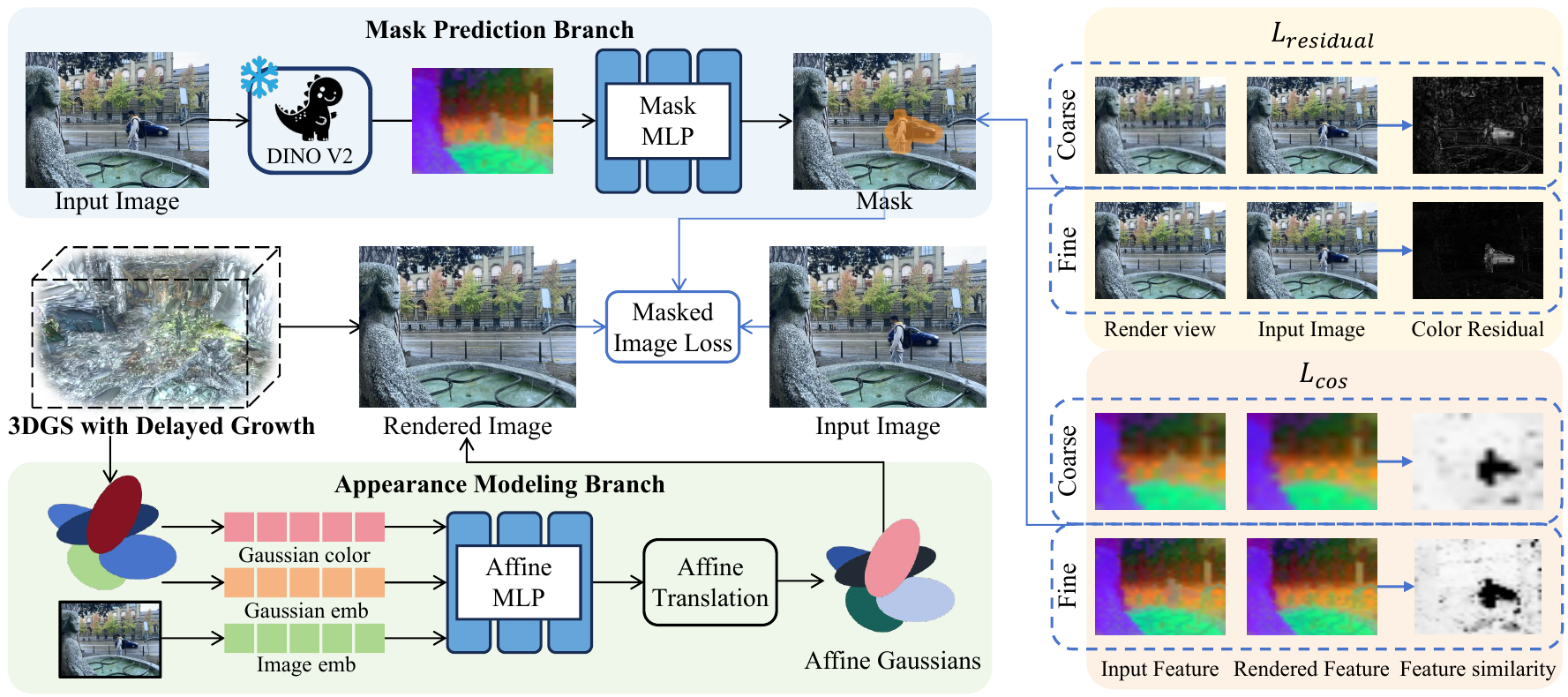}
    \caption{Overview of the proposed method. The main reconstruction pipeline employs 3DGS with \textbf{\DelayedGS}, generating rendered images that are optimized with a masked reconstruction loss. The method handles two types of in-the-wild inputs:
    (1) Images with transient distractors, where a Mask Prediction Branch predicts per-pixel masks to guide transient suppression, with the masks supervised by \textbf{\CoarseToFine};
    (2) Images with both transients and illumination variations, where an Appearance Modeling Branch predicts affine coefficients from 2D, 3D embedding, and original Gaussian colors to modulate the affine Gaussian colors.}
    \label{fig:pipeline}
\end{figure*}

\paragraph{Robustness to Transient Distractors in 3DGS}
Unlike NeRF, which uses a continuous MLP-based implicit representation, 3DGS employs a discrete explicit representation.  
For distractor removal, transient objects are typically filtered out using masks~\cite{xu2024wild, ungermann2024robust,wang2024distractor,xu2024splatfacto,dahmani2024swag,wang2024uw,bao2024distractor}.

To handle transient objects, WildGaussians~\cite{kulhanek2024wildgaussians} incorporates the DINO~\cite{oquab2023dinov2} features to predict uncertainty, which is then converted into a mask.
Robust3DGaussians~\cite{ungermann2024robust} enhances the predicted mask by leveraging SAM~\cite{kirillov2023segment}.
SpotLessSplats~\cite{sabour2024spotlesssplats} leverages features from Stable Diffusion~\cite{rombach2022high}, designing two clustering strategies for mask prediction. 
T-3DGS~\cite{pryadilshchikov2024t} introduces an unsupervised transient detector based on a consistency loss and a video object segmentation module to track objects in the videos.

More recently, DeSplat~\cite{wang2024desplat} decomposes the 3DGS scenes into a static 3DGS and per-view transient 3DGS by only minimizing the photometric loss. HybridGS~\cite{lin2024hybridgs} instead combines 3DGS with per-view 2D image Gaussians to decouple dynamics and statics.
{DeGauss~\cite{wang2025degauss} adopts the decoupled foreground-background design, which leverages the expressiveness of 3DGS static background and 4DGS dynamic foreground.}
DAS3R~\cite{xu2024das3r} and RoMo~\cite{goli2024romo} proposed to estimate motion mask for dynamic videos by making use of the temporal consistency constraints, which cannot be directly applied to a set of unordered images.
R3GS~\cite{wang2025r3gs} fine-tunes a detection model and incorporates sky segmentation priors to predict transient masks.
AsymGS~\cite{li2025asymgs} integrates binary masks obtained from SAM and multiple heuristic cues with complementary feature-based probabilistic masks to eliminate transient distractors.
Different from existing methods, we analyze the densification process of 3DGS and propose a simple yet effective solution based on the \delayedGS and \coarsetofine to reliably remove the effects of transient objects.

\paragraph{Robustness to Illumination Variations in 3DGS} 
To handle illumination variations in in-the-wild scenes, many studies~\cite{zhang2024gaussian, wang2024we, darmon2024robust, kulhanek2024wildgaussians,tang2024nexussplats,li2025asymgs} consider transient distractions while incorporating appearance modeling, and explore strategies that combine the global information of reference images with local 3D features. 

SWAG~\cite{dahmani2024swag} predicts affine Gaussian color for appearance variation and opacity variation for transient objects using 2D image embeddings and hash–based 3D embeddings.
WildGaussians~\cite{kulhanek2024wildgaussians} leverages 2D image embeddings and per-Gaussian embeddings to obtain the color affine transformation matrix, which adjusts the Gaussian SH color to match the style of a reference image.
Splatfacto-W~\cite{xu2024splatfacto} considers the infinite-distance background prior and implicitly decomposes the background Gaussians for separate modeling. 
More recently, Look at the Sky~\cite{wang2025look} decouples the sky region using a pre-trained segmentation model and introduces an implicit cubemap, which models the foreground and background separately.
R3GS~\cite{wang2025r3gs} combines CNN-based image embeddings and hash-based 3D embeddings, and decouples Gaussian properties through multiple MLP networks.
GaRe~\cite{bai2025gare} focuses on rendering under complex illumination variations without considering transient effects, by decomposing global illumination into multiple physically interpretable components and modeling them independently or collaboratively.
In this paper, we focus on analyzing the potential impact of delayed Gaussian growth on appearance modeling and extend our transient-free 3DGS to handling illumination challenges.

\paragraph{Optimization in Densification and Regularization} 
There are prior works aiming to improve the densification and optimization process of 3DGS~\cite{zhang2024fregs,fang2024mini,bulo2024revising,hyung2024effective,bulo2024revising}.
For example, several methods~\cite{ye2024absgs, zhang2024pixel,yu2024gaussian} have analyzed the gradient computation process and identified issues such as gradient collision or averaging, which lead to suboptimal reconstruction quality. 
RAIN-GS~\cite{jung2024relaxing} investigates an alternative initialization strategy for 3DGS without relying on Colmap SFM.
These methods do not consider the effect of transient objects.
In this work, we analyze and leverage the behaviors of Gaussian densification in the context of transient-free 3D reconstruction.

\section{Transient-free 3DGS Optimization}
\label{sec:method}

In this section, we introduce our solution for handling transient distractors. We will extend this solution for handling illumination variations in \sref{sec:method_illumination}.

\subsection{Overview}
\label{sub:Overview}
Given casually captured multi-view posed images with transient objects, our goal is to optimize a clean 3D Gaussian splatting representation that enables distractor-free novel-view synthesis.
Our approach builds upon recent robust 3DGS methods that jointly optimize 3D representation and transient object masks during training~\cite{sabour2024spotlesssplats}.
The transient masks selectively filter dynamic regions in images, while improving scene modeling by providing more accurate supervision for mask MLP optimization. 

However, this interdependence can lead to instability in early training.
On the one hand, if the masks are too small, they fail to filter all transient regions, causing newly generated Gaussians to fit transient objects. This makes it difficult to remove artifacts in later stages. 
On the other hand, the static scene reconstruction is often overly smooth in the early stage, which will misguide the mask MLP into incorrectly classifying static regions as dynamic, hindering their reconstruction and leading to under-representation of static content.

To address these challenges, we introduce two effective designs (see~\fref{fig:pipeline}).
First, we introduce a \delayedGS strategy to postpone the Gaussian densification process to prevent fitting transient objects in the early stage. 
Second, we propose a \coarsetofine approach to refine mask predictions over time, reducing the misclassification of static regions as transient and improving the optimization of static content.

\subsection{3DGS with Transient Mask Estimation}
\paragraph{3D Gaussian Splatting} 
We represent the scene as a set of 3D Gaussians $\model{=} \{ \element_i \}_{i=1}^N$. Each Gaussian color is represented using spherical harmonics (SH) to model view-dependent color~\cite{kerbl20233d}. 
For novel view synthesis, 3D Gaussians are projected to 2D and rendered by differentiable rasterization using alpha blending~\cite{zwicker2001surface}. The final pixel color $\gcolor(p)$ is computed via alpha blending:
\begin{equation}
\gcolor(p) = \sum^N_{i=1} \gcolor_i\,\alpha_i\,\model^{2D}_i \prod_{j=1}^{i-1} (1 - \alpha_j\,\model^{2D}_j),
\end{equation}
where $\gcolor_i$ is the color computed from spherical harmonics coefficients with the view direction, $\opacity_i$ is the opacity, $\model^{2D}_i$ is the 2D projection of the $i$-th Gaussian.

The 3DGS is optimized by minimizing the L1 loss and SSIM loss between the rendered and the captured images:
\begin{equation}
\loss = ( 1 - \lambda ) \loss_1 + \lambda \loss_\textrm{D-SSIM}.
\end{equation}
During optimization, adaptive density control periodically clones/prunes Gaussians based on accumulated positional gradient magnitudes.

\begin{figure}[t] \centering
    \makebox[0.116\textwidth]{\footnotesize }
    \makebox[0.116\textwidth]{\footnotesize DINOv2}
    \makebox[0.116\textwidth]{\footnotesize SAM2}
    \makebox[0.110\textwidth]{\footnotesize StableDiffusion}\\
    \vspace{1pt}
    \includegraphics[width=0.48\textwidth]{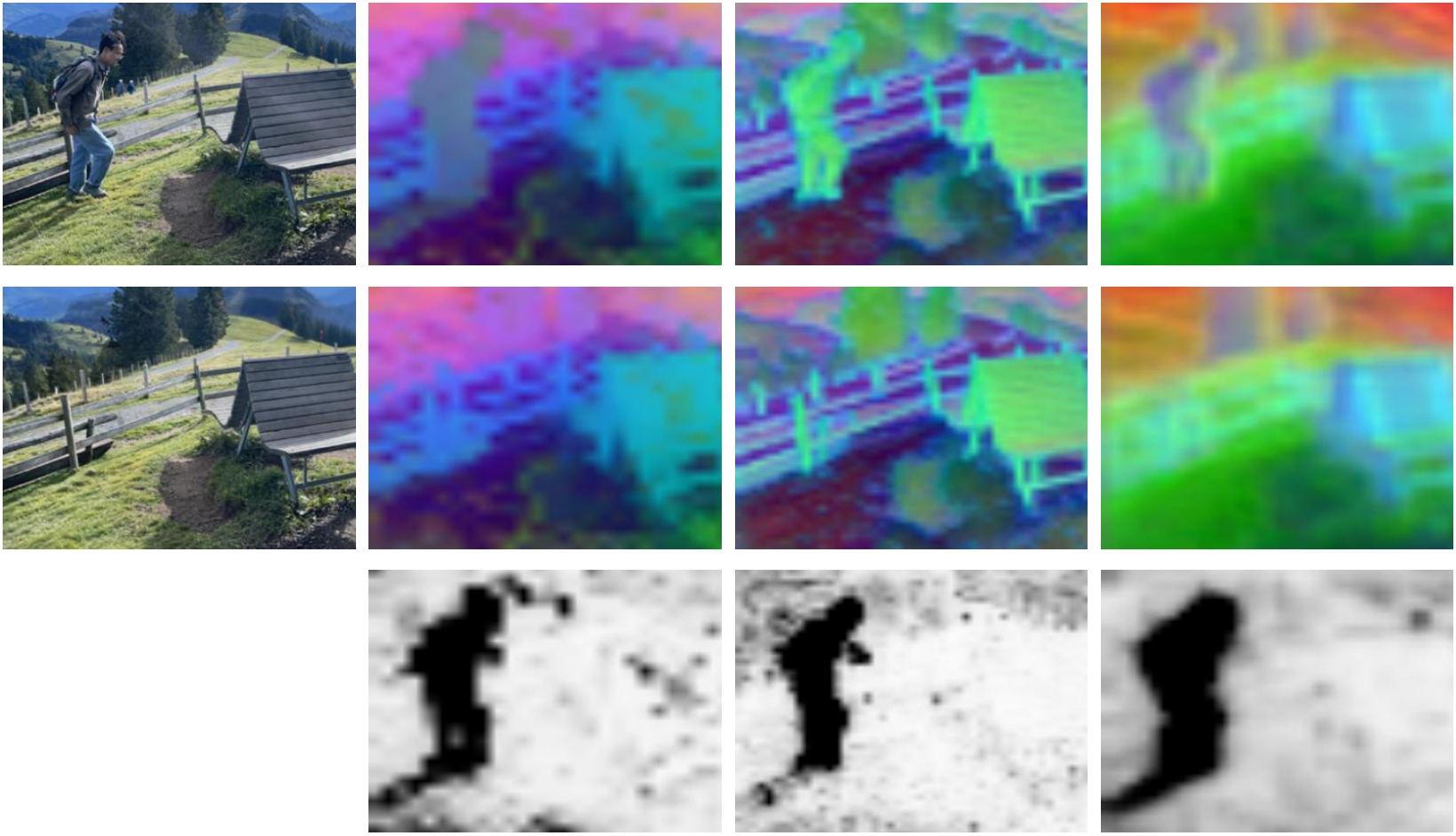}
    \caption{Visualization of DINOv2, SAM2, and SD features via PCA. The last row compares the cosine similarity maps between features of the ground-truth and rendered image.} \label{fig:feat_compare}
\end{figure}

\paragraph{Transient Mask Estimation}
To deal with transient objects, following recent work~\cite{sabour2024spotlesssplats,kulhanek2024wildgaussians}, we predict per-image transient masks $M_t$ using an MLP conditioned on image features $f_t$:
\begin{equation}
M_t = \textrm{Sigmoid}(\textrm{MLP}_\textrm{mask}(f_t)).
\end{equation}

The estimated mask is then used to apply a masked photometric loss that excludes transient regions.

Recent works utilize features containing strong semantic information as MLP inputs (e.g., DINOv2~\cite{oquab2023dinov2,kulhanek2024wildgaussians}, StableDiffusion~\cite{rombach2022high,sabour2024spotlesssplats}, SAM~\cite{kirillov2023segment,goli2024romo}). 
Our preliminary experiments found that StableDiffusion features provide stronger semantic information, but they are computationally expensive to extract.
Although SAM features are better at producing mask with more accurate boundary, they struggle to locate the shadow regions, which produce incomplete mask prediction, as shown in~\fref{fig:feat_compare}.
We employ the DINOv2 features as input to the MLP as they maintain a good balance of computational efficiency and semantic extraction ability.

\paragraph{Optimization of Mask MLP}
The optimization of the MLP weights requires appropriate supervision. 
We adopt the image robust loss $\mathcal{L}_\textrm{residual}$ based on the image residual information introduced in \cite{sabour2024spotlesssplats} as a form of supervision.

In addition, to better leverage deep high-dimensional feature information extracted from images, which have different properties from the image residual, we adopt a feature robust loss $\mathcal{L}_\textrm{cos}$ utilizing the information of feature similarity between the rendered and captured images.
Specifically, we extract DINOv2 features of the real image $f_t$ and rendered image $f_t'$, and compute their cosine similarity map. 
Then we convert the cosine similarity map to be in the value range of $[0, 1]$ following \cite{kulhanek2024wildgaussians}:
\begin{align}
    \label{eq:}
    M_\textrm{cos} = \textrm{max} \left( 2\cos\left ( f_t,f_t' \right ) -1,0 \right),
\end{align}
where $M_\textrm{cos}$ equals $1$ if the feature cosine similarity is $1$, and equals $0$ if the similarity is less than $0.5$.
Then the feature robust loss is expressed as:
\begin{equation}
    \mathcal{L}_\textrm{cos} = \left \| M_t -  M_\textrm{cos} \right \|_1.
\end{equation}

The MLP is optimized using the following loss:
\begin{equation}
    \mathcal{L}_\textrm{MLP} = \lambda_\textrm{residual} \mathcal{L}_\textrm{residual} + \lambda_\textrm{cos} \mathcal{L}_\textrm{cos},
\end{equation}
where $\lambda_\textrm{residual}$, $\lambda_\textrm{cos}$ are the corresponding weights for image robust supervision and feature robust loss, respectively.

\subsection{Delayed Gaussian Growth for Mask Learning}

Motivated by our observation that disabling Gaussian densification in 3DGS significantly improves the learning of low-frequency static components, we introduce a delayed Gaussian growth strategy, modifying 3DGS~\cite{kerbl20233d} to defer Gaussian densification during optimization.

\begin{figure}[tb] \centering
    \includegraphics[width=0.24\textwidth]{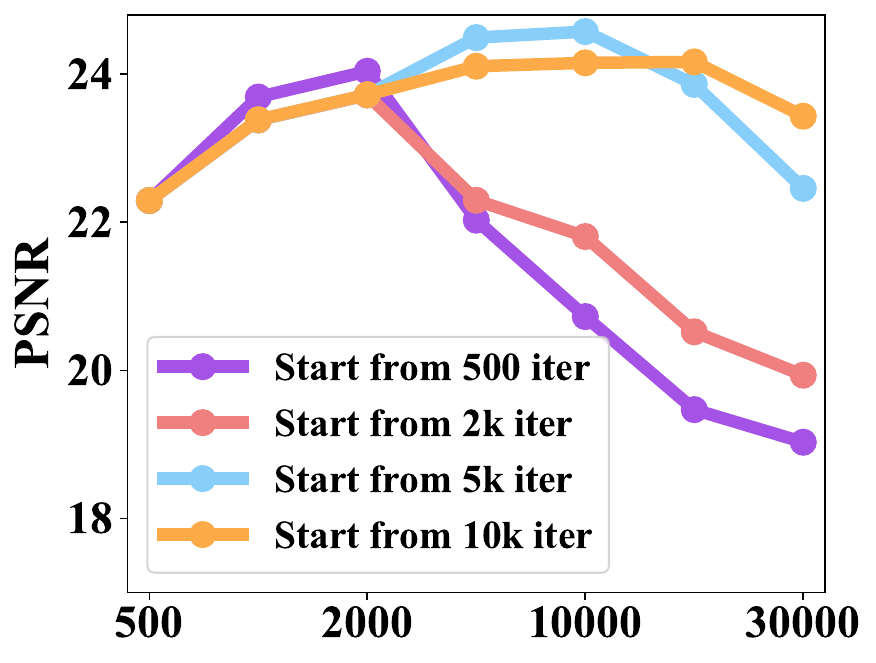}
    \includegraphics[width=0.24\textwidth]{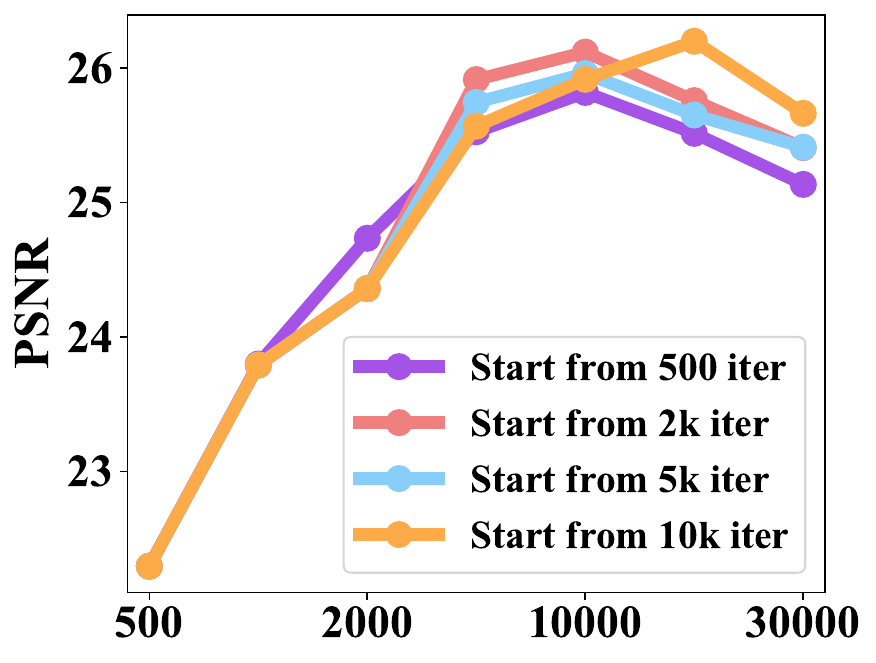}
    \\
    \vspace{-6pt}
    \makebox[0.24\textwidth]{\small (a) w/o robust mask} 
    \makebox[0.24\textwidth]{\small (b) with robust mask} 
    \caption{Effects of start iteration of Gaussian densification with and without the transient mask learning. } \label{fig:analysis_start}
\end{figure}

\paragraph{Analysis of Delayed Gaussian Growth} 
To evaluate the impact of the Gaussian densification start time in 3DGS, we vary the initial densification iteration while keeping the densification interval fixed at 10K iterations. 
As shown in~\fref{fig:analysis_start}~(a), delaying densification allows 3DGS to focus on reconstructing the static scene during the early training stages. 
However, once densification begins, newly introduced Gaussians tend to fit transient objects, leading to a decline in PSNR metrics. Notably, models with earlier densification exhibit worse performance, indicating that premature densification promotes transient object fitting. 
These results suggest that postponing densification helps the model better capture the static components before accommodating dynamic elements.

\paragraph{Mask Learning with Delayed Gaussian Growth} 
To mitigate transient artifacts caused by uncontrolled Gaussian growth, we incorporate transient mask learning into the delayed densification process. 
As shown in~\fref{fig:analysis_start}~(b), this integration significantly improves reconstruction accuracy by leveraging mask predictions to regulate Gaussian expansion.
By leveraging mask predictions to regulate Gaussian expansion, this approach effectively suppresses transient artifacts and enhances scene fidelity.
In particular, variants with a later densification start achieve more accurate results. These results demonstrate that transient mask learning and delayed densification work collaboratively to enhance the stability and accuracy of 3DGS optimization.

\paragraph{Mask Regularization at Early Stage} 
The timing of applying transient mask filtering in 3DGS is a critical aspect. In the initial training phase, rendered images exhibit low quality with large image residuals and poor feature similarity, leading to inaccurate mask estimation. 
To mitigate this, prior methods either delay mask learning until after a warm-up period (e.g., 1500 iterations)~\cite{kulhanek2024wildgaussians} or employ random mask sampling strategies~\cite{sabour2024spotlesssplats}. 
However, delaying mask application risks incorporating transient objects into the scene, making them harder to remove later.

Thanks to the delayed strategy for Gaussian growth, our approach ensures that early-stage optimization focuses solely on static scenes.
To facilitate the optimization of static regions across the entire scene, we encourage the mask MLP to initially classify all regions as static and gradually filter out transient objects.
To achieve this, we introduce a regularization term into the mask MLP’s supervision:
\begin{equation}
    \loss_\textrm{reg} = \exp{(-\frac{i}{\beta_\textrm{reg}})}\left \| 1 - M_t \right \|_1,
\end{equation}
where $i$ is the iteration number of training,  
$\left \| 1 - M_t \right \|_1$ is weighted by $\exp{(-i/\beta_\textrm{reg})}$, the entire regularization term equals $1$ when $i$ is $0$ and decays as $i$ increases.

The overall loss for mask optimization is expressed as:
\begin{equation}
    \mathcal{L}_\textrm{MLP} = \lambda_\textrm{residual} \mathcal{L}_\textrm{residual} + \lambda_\textrm{cos} \mathcal{L}_\textrm{cos} + \lambda_\textrm{reg} \mathcal{L}_\textrm{reg},
    \label{eq:loss_mask}
\end{equation}
where $\lambda_\textrm{reg}$ is the corresponding weight for regularization.

\begin{figure}[t] \centering
    \makebox[0.003\textwidth]{\scriptsize}
    \makebox[0.088\textwidth]{\scriptsize Masked GT}
    \makebox[0.088\textwidth]{\scriptsize Rendering}
    \makebox[0.082\textwidth]{\scriptsize GT Feature}
    \makebox[0.088\textwidth]{\scriptsize Rendered Feature}
    \makebox[0.08\textwidth]{\scriptsize Cosine}
    \includegraphics[width=0.48\textwidth]{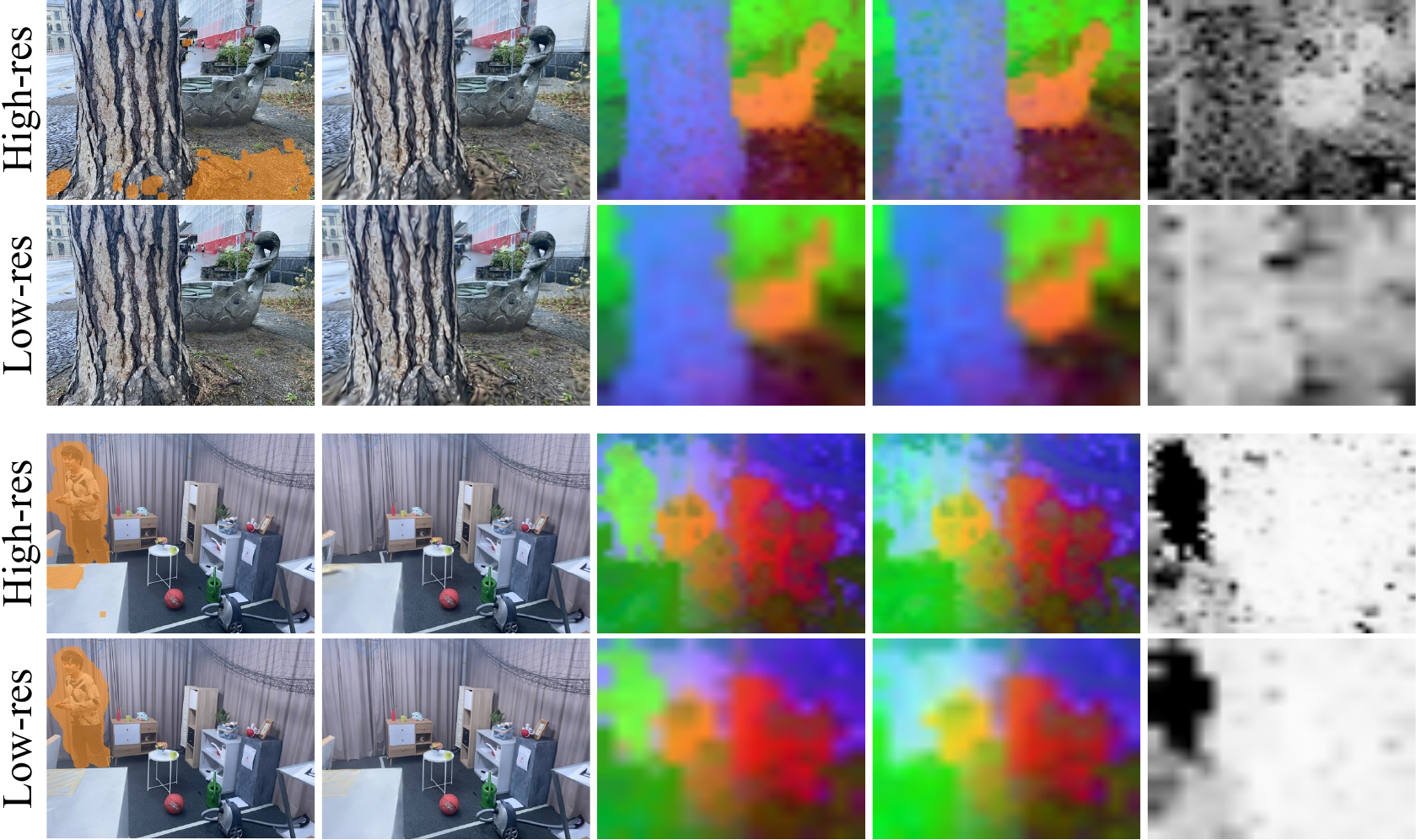}
    \caption{Effects of mask supervisions derived from different resolutions on two scenes. The first column shows input images overlapped with yellow masks predicted by the mask MLP after training with supervisions derived from the corresponding resolutions.} \label{fig:analysis_mask_feature}
\end{figure}

\subsection{Scaled-cascaded Mask Bootstrapping}

While our \delayedGS strategy effectively mitigates the influence of transient regions by focusing optimization on static areas, the under-reconstruction of static scenes remains an issue in the early stages. This problem arises due to the sparsity of the initial Gaussian points, particularly in large-scale unbounded outdoor scenes. 
Consequently, the rendered outputs in these regions appear overly smooth, leading to large image residuals and low feature similarity. This, in turn, causes the mask MLP to misclassify under-reconstructed static areas as dynamic.

\paragraph{Robust Feature Similarity Computation}
To address this, we aim to make the supervision signal more tolerant to under-reconstructed regions in the early optimization phase. We observe that while high-resolution features extracted from high-resolution images provide fine-grained spatial details, they suffer from limited receptive fields and increased sensitivity to local noise. In contrast, low-resolution features capture global consistency more effectively, as each patch integrates broader contextual information, inherently suppressing local noise in feature representations.

As shown in~\fref{fig:analysis_mask_feature}, compared to high-resolution images, low-resolution images naturally suppress fine details, leading to smoother color residuals and feature similarity. 
This suggests that evaluating residuals and feature similarity at a lower resolution during the early stages improves robustness, as it allows under-reconstructed static regions to be retained while maintaining sensitivity to transient areas.

\paragraph{Coarse-to-fine Mask Supervision}
Building on this insight, we propose a resolution-cascaded approach that progressively refines mask supervision by transitioning from low-resolution to high-resolution signals. This method helps the mask MLP retain more static regions in the early optimization phase.

Specifically, before the start of Gaussian densification,  we render low-resolution images from 3DGS to compute low-resolution image residuals and feature consistency to supervise the mask MLP. 
Once densification begins, we switch to high-resolution residuals and cosine similarity between high-resolution features, ensuring finer-grained discrimination of transient and static regions.

\begin{figure*}[ht] 
\begin{center}
    \includegraphics[width=\textwidth]{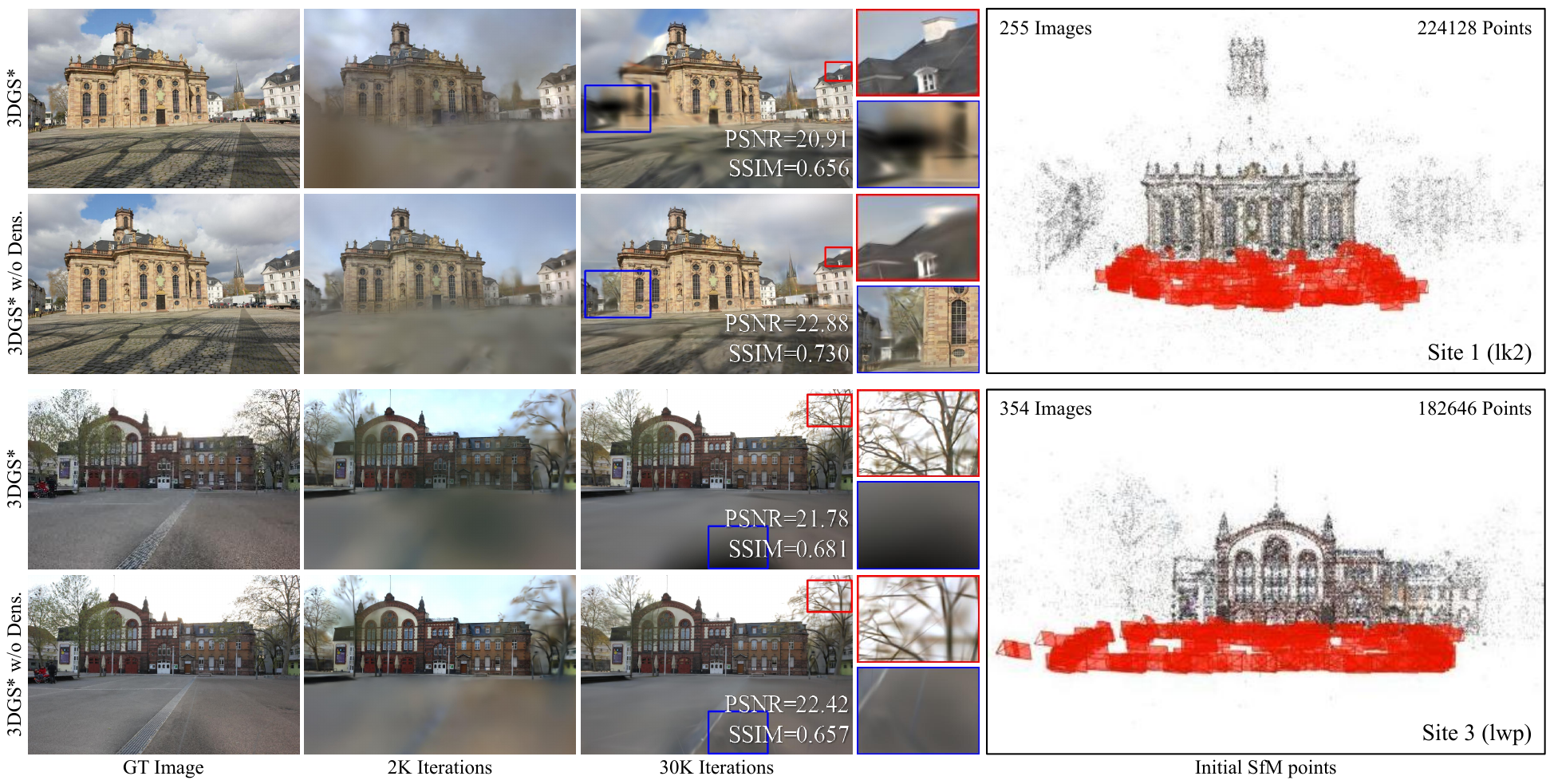}\\
    \vspace{-2pt}
    \caption{{\textbf{Analysis of Gaussian densification in illumination variation fitting.} 
    3DGS* builds upon 3DGS~\cite{kerbl20233d} by incorporating the appearance modeling introduced in \sref{sec:method_illumination} and using ground-truth masks to eliminate transient effects.
    As training progresses, 3DGS suffers from performance degradation and exhibits artifacts due to the increasing number of Gaussians. Disabling Gaussian densification notably improves the quality of rendering. 
    Despite eliminating the floaters (zoom-in by \textcolor{blue}{blue box}), \emph{3DGS* w/o densification} struggles to recover fine details in regions with sparse Gaussian initialization (zoom-in by \textcolor{red}{red box}).}}
    \label{fig:dens_illumination}
\end{center} 
\end{figure*}

\begin{table*}[t] \centering
    \begin{center}
    \captionof{table}{Quantitative comparison on NeRF On-the-go dataset~\cite{ren2024nerf}. The best results are highlighted in \textbf{bold}, and the second in \underline{underline}.}
    \label{table:onthego}
    \resizebox{\textwidth}{!}{
    \begin{tabular}{l|ccc|ccc|ccc|ccc|ccc|ccc}
    \toprule
    \multirow{3}{*}{Method} 
    & \multicolumn{6}{c|}{Low Occlusion}  
    & \multicolumn{6}{c|}{Medium Occlusion} 
    & \multicolumn{6}{c}{High Occlusion} 
\\
    \multicolumn{1}{l|}{}  
    & \multicolumn{3}{c}{Mountain}  
    & \multicolumn{3}{c|}{Fountain} 
    & \multicolumn{3}{c}{Corner}  
    & \multicolumn{3}{c|}{Patio} 
    & \multicolumn{3}{c}{Spot} 
    & \multicolumn{3}{c}{Patio-High}  
    \\
     & PSNR & SSIM & LPIPS 
     & PSNR & SSIM & LPIPS 
     & PSNR & SSIM & LPIPS 
     & PSNR & SSIM & LPIPS 
     & PSNR & SSIM & LPIPS 
     & PSNR & SSIM & LPIPS 
\\
    \midrule
    3DGS~\cite{kerbl20233d}
    & 19.21 & 0.691 & 0.229 
    & 20.08 & \underline{0.686} & \underline{0.208} 
    & 22.65 & 0.835 & 0.162 
    & 17.04 & 0.713 & 0.232 
    & 18.54 & 0.717 & 0.334 
    & 17.04 & 0.657 & 0.314 
\\
    SpotLessSplat~\cite{sabour2024spotlesssplats}
    & 20.67 & 0.670 & 0.282 
    & \underline{20.63} & 0.645 & 0.265 
    & 25.47 & 0.858 & 0.155 
    & \underline{21.43} & 0.803 & 0.171 
    & 23.64 & 0.819 & 0.207 
    & 21.17 & 0.749 & 0.237 
\\
    WildGaussians~\cite{kulhanek2024wildgaussians}
    & \underline{20.77} & 0.697 & 0.268 
    & 20.48 & 0.666 & 0.250 
    & 25.21 & 0.865 & 0.136 
    & 21.17 & \underline{0.804} & \underline{0.168} 
    & 24.60 & 0.871 & 0.135 
    & \underline{22.44} & 0.802 & 0.184
\\
    Robust3DGaussians~\cite{ungermann2024robust}
    & 19.47 & 0.672 & 0.251 
    & 19.74 & 0.653 & 0.254 
    & 24.41 & 0.869 & 0.118 
    & 16.63 & 0.729 & 0.209 
    & 22.64 & 0.874 & 0.132 
    & 21.56 & 0.799 & 0.174 
\\
    DeSplat~\cite{wang2024desplat}
    & 19.58 & \underline{0.702} & \textbf{0.201} 
    & 20.39 & 0.675 & 0.211 
    & \underline{25.93} & \underline{0.874} & \underline{0.116} 
    & 18.71 & 0.775 & 0.179 
    & \underline{25.65} & \underline{0.887} & \underline{0.125} 
    & 22.11 & \underline{0.809} & \underline{0.169} 
 \\
    RobustSplat 
    & \textbf{21.15} & \textbf{0.737} & \textbf{0.201} 
    & \textbf{21.01} & \textbf{0.701} & \textbf{0.199} 
    & \textbf{26.42} & \textbf{0.897} & \textbf{0.104} 
    & \textbf{21.63} & \textbf{0.827} & \textbf{0.139} 
    & \textbf{26.21} & \textbf{0.907} & \textbf{0.102} 
    & \textbf{22.87} & \textbf{0.837} & \textbf{0.146} 
\\
    \bottomrule
    \end{tabular}}

    \end{center}
    \vspace{-2pt}
    \makebox[0.195\textwidth]{\footnotesize WildGaussians~\cite{kulhanek2024wildgaussians}}
\makebox[0.195\textwidth]{\footnotesize SpotLessSplats~\cite{sabour2024spotlesssplats}}
\makebox[0.195\textwidth]{\footnotesize DeSplat~\cite{wang2024desplat}}
\makebox[0.195\textwidth]{\footnotesize RobustSplat}
\makebox[0.195\textwidth]{\footnotesize GT}
\\
\includegraphics[width=0.195\textwidth]{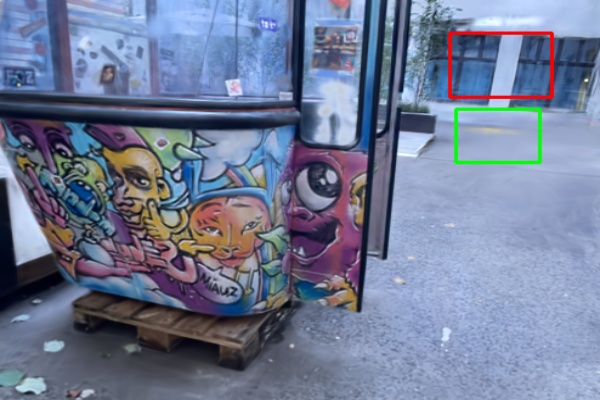}
\includegraphics[width=0.195\textwidth]{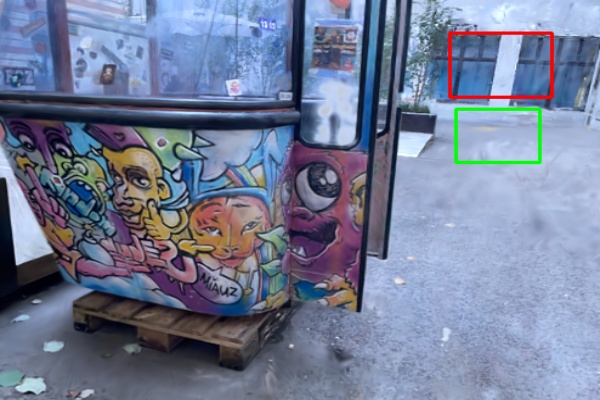}
\includegraphics[width=0.195\textwidth]{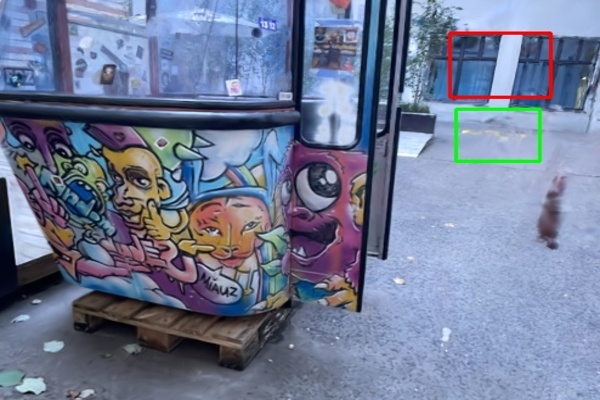}
\includegraphics[width=0.195\textwidth]{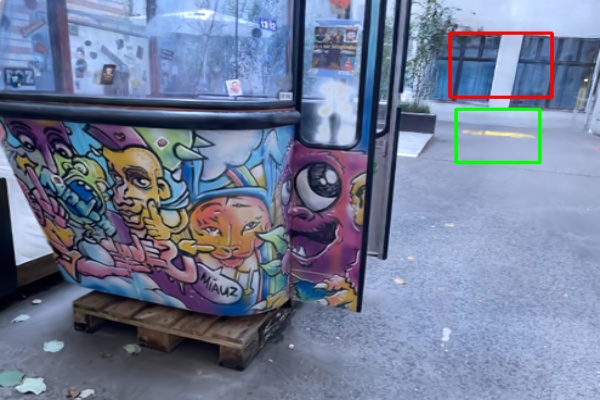}
\includegraphics[width=0.195\textwidth]{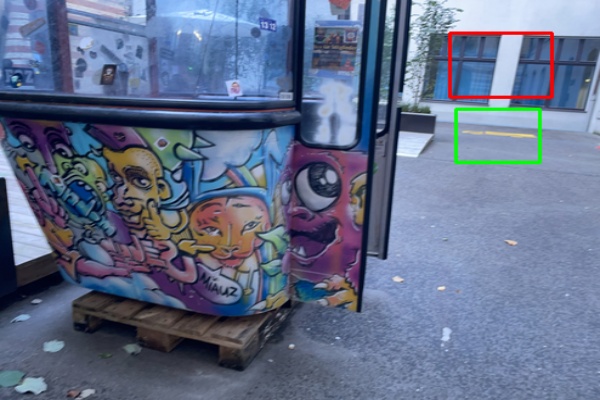}
\\
\includegraphics[width=0.0948\textwidth,height=0.07\textwidth]{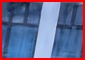}
\includegraphics[width=0.0948\textwidth,height=0.07\textwidth]{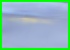}
\includegraphics[width=0.0948\textwidth,height=0.07\textwidth]{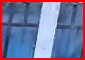}
\includegraphics[width=0.0948\textwidth,height=0.07\textwidth]{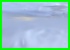}
\includegraphics[width=0.0948\textwidth,height=0.07\textwidth]{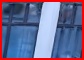}
\includegraphics[width=0.0948\textwidth,height=0.07\textwidth]{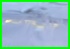}
\includegraphics[width=0.0948\textwidth,height=0.07\textwidth]{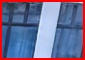}
\includegraphics[width=0.0948\textwidth,height=0.07\textwidth]{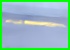}
\includegraphics[width=0.0948\textwidth,height=0.07\textwidth]{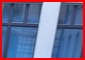}
\includegraphics[width=0.0948\textwidth,height=0.07\textwidth]{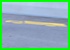}
\\[0.5em]
\includegraphics[width=0.195\textwidth]{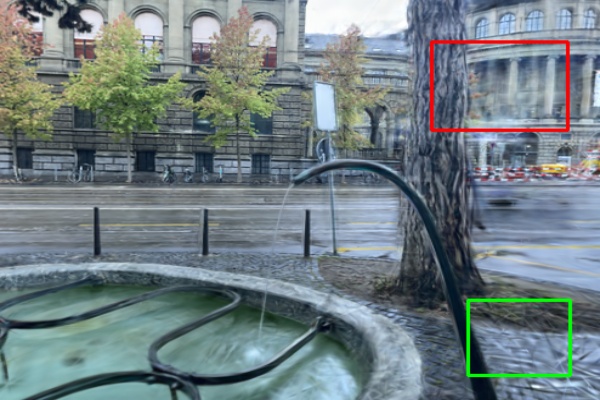}
\includegraphics[width=0.195\textwidth]{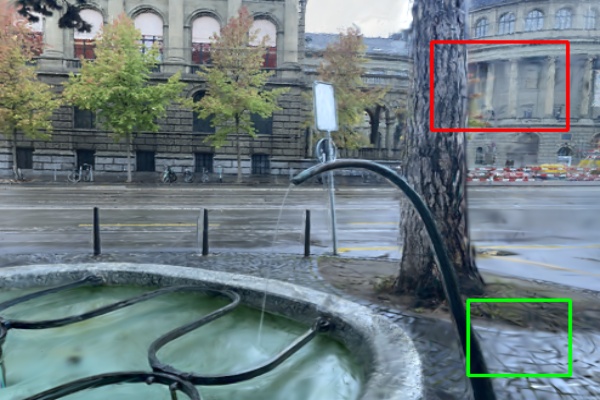}
\includegraphics[width=0.195\textwidth]{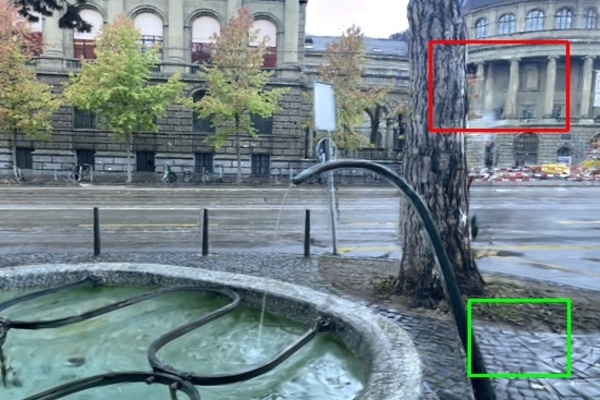}
\includegraphics[width=0.195\textwidth]{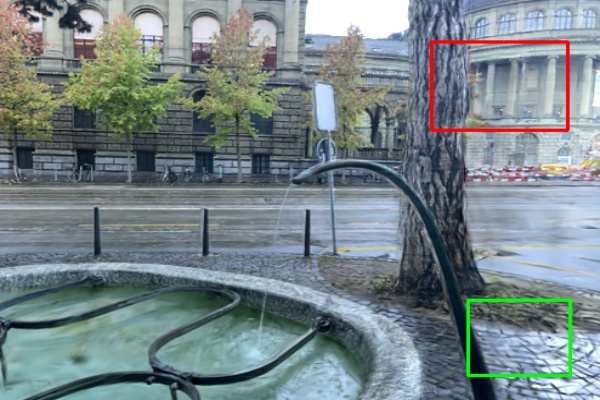}
\includegraphics[width=0.195\textwidth]{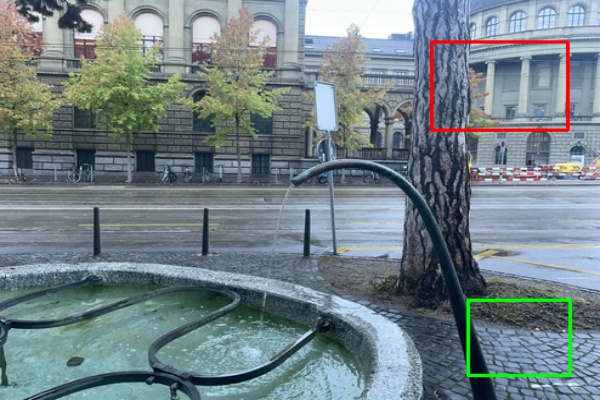}
\\
\includegraphics[width=0.0948\textwidth,height=0.07\textwidth]{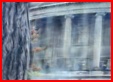}
\includegraphics[width=0.0948\textwidth,height=0.07\textwidth]{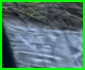}
\includegraphics[width=0.0948\textwidth,height=0.07\textwidth]{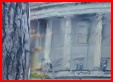}
\includegraphics[width=0.0948\textwidth,height=0.07\textwidth]{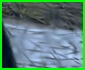}
\includegraphics[width=0.0948\textwidth,height=0.07\textwidth]{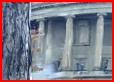}
\includegraphics[width=0.0948\textwidth,height=0.07\textwidth]{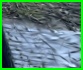}
\includegraphics[width=0.0948\textwidth,height=0.07\textwidth]{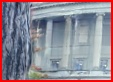}
\includegraphics[width=0.0948\textwidth,height=0.07\textwidth]{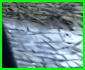}
\includegraphics[width=0.0948\textwidth,height=0.07\textwidth]{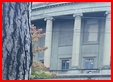}
\includegraphics[width=0.0948\textwidth,height=0.07\textwidth]{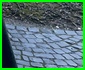}
\\

    \captionof{figure}{Qualitative results on \emph{Patio-high} and \emph{Fountain} from NeRF On-the-go dataset~\cite{ren2024nerf}.}
    \label{fig:qual_onthego}
    \vspace{-1.2em}
\end{table*}

\section{Illumination-robust 3DGS Optimization}
\label{sec:method_illumination}

In this section, we first analyze the challenges of illumination variations and discuss how the Gaussian optimization process is affected by illumination variations. Then, we provide our solution that incorporates the \delayedGS with appearance modeling.

\subsection{Analysis for Illumination Variations}
To model appearance variations caused by lighting, prior work~\cite{lin2024vastgaussian} uses 2D image embeddings to perform color adjustment. However, this approach is limited to global color shifts and cannot capture local effects such as reflections or shadows. Recent methods~\cite{martin2021nerf,kulhanek2024wildgaussians} address this by incorporating 3D features, following a joint 2D and 3D embedding paradigm to handle complex appearance. However, scenes with significant illumination variations still exhibit artifacts.
We identify that the Gaussian densification process can potentially introduce the artifacts.

To illustrate how Gaussian optimization impacts appearance modeling capabilities, we analyze the relationship between densification and appearance modeling on NeRF-OSR dataset~\cite{rudnev2022nerfosr} that contains scene images captured under different periods and seasons. We incorporate 3DGS with appearance modeling, which will be introduced in \sref{sec:appearance_modeling}, and compare the models with and without Gaussian densification. To eliminate the effect of transient distractors, we use the ground-truth masks provided by \cite{rudnev2022nerfosr}, with analysis results shown in \fref{fig:dens_illumination}. 

\paragraph{Observation}
We find that densification in 3DGS tends to introduce redundant points in underconstrained regions, leading to floaters. Disabling the densification process in 3DGS avoids the issues and maintains global consistency, sacrificing only marginal high-frequency details. 

The main reason is that in the early stage of 3DGS training, the scene representation remains under-reconstructed. The per-image 2D embeddings and 3D embeddings used to model illumination variation have not yet converged, especially when the training set contains many images. This causes the photometric reconstruction loss to be significantly larger. Unfortunately, this loss value alone cannot distinguish whether a large error stems from under-reconstruction or from unmodeled lighting changes. As a result, the position gradients required for densification are more easily triggered in vanilla 3DGS, leading to abnormal Gaussian growth. Furthermore, appearance modeling tends to force intermediate renderings toward the ground-truth appearance, inadvertently disrupting the optimization.

This observation reveals that our delayed Gaussian growth strategy, originally designed to handle transients, notably suits scenes with complex illumination without requiring any additional design.

\subsection{Appearance Modeling}
\label{sec:appearance_modeling}
To handle varying illumination, a simple idea is to design a learnable color affine matrix, which models the mapping from global color to per-image color. Vanilla 3DGS~\cite{kerbl20233d} adopts this design, referred to as 3DGS-E\footnote{3DGS-E is the exposure-compensated extension of 3DGS~\cite{kerbl20233d} in official implementation: \href{https://github.com/graphdeco-inria/gaussian-splatting}{https://github.com/graphdeco-inria/gaussian-splatting}.}. 
However, this learnable matrix is constrained by its limited parameters, which makes it difficult to capture complex appearance changes.

\paragraph{Color Affine Transformation}
To model complex illumination variations, existing methods~\cite{kulhanek2024wildgaussians,wang2024desplat} follow a paradigm that combines 2D embeddings with 3D embeddings, aiming to separately capture global appearance changes and local brightness or color variations caused by sunlight and lamplight. Specifically, the affine coefficients $\alpha$ and $\beta$ of color transformation are predicted by an MLP, which is conditioned on the 2D image embedding $f_\textrm{img}$, the 3D Gaussian embedding $f_\textrm{gs}$, and the Gaussian color $\gcolor_i$:
\begin{equation}
    \left(\alpha, \beta\right) = \mathrm{MLP_{affine}}\left(\gcolor_i,f_\textrm{img}, f_\textrm{gs}\right), 
\end{equation}
where $\gcolor_i$ is the color computed from spherical harmonics coefficients with the view direction, $\alpha$ is the scale factor for color translation, and $\beta$ is the color translation bias. Then the affine Gaussian SH color $\gcolor_i^\textrm{aff}$ is expressed as:
\begin{equation}
    \gcolor_i^\mathrm{aff} = \alpha \odot \gcolor_i + \beta .
\end{equation}

\paragraph{Appearance Modeling with Delayed Gaussian Growth} 
Based on our analysis, we find that the premature densification designed for static reconstruction introduces artifacts that are difficult to eliminate. To mitigate these artifacts, our \delayedGS for handling transients is potentially suitable for this challenge. The main reason is that it tolerances the fitting of appearance in under-reconstructed regions, allowing the representation to prioritize reconstructing the geometry.

Fixing the Gaussian count in early stages enables stable convergence of 2D image embeddings by eliminating gradient noise introduced during densification. 
In addition, the correlation between each 3D Gaussian embedding and the total Gaussian count allows delayed Gaussian growth to effectively enforce a coarse-to-fine learning regime. As a result, the model first captures coarse color transformations prior to densification and incrementally refines them throughout the densification process.
The ablation results in \sref{sec:ablation_illumination} demonstrate the effectiveness of this strategy, indicating that it contributes to improving reconstruction quality.

\paragraph{Integration with RobustSplat} 
Through Appearance Modeling, the appearance changes of any unconstrained data can be simulated. We incorporate it into \emph{RobustSplat} to make up for the deficiencies of RobustSplat in scenes with illumination changes. In particular, we have modified the loss function for image rendering to regularize affine colors~\cite{kulhanek2024wildgaussians}:
\begin{equation}
\mathcal{L} = (1-\lambda ) M \cdot \| C^\textrm{aff} - C^\textrm{gt} \|_1  + \lambda M \cdot \textrm{D-SSIM}(C, C^\textrm{gt}),
\end{equation}
where $C^\textrm{aff}$ is the affine rendered image, $C$ is the raw rendered image, and $C^\textrm{gt}$ is the ground-truth image. 

The unconverged color transformation at early stage would affect both the color residual and the feature cosine similarity for mask predicted supervision. To minimize its impact on mask prediction, we modify \eref{eq:loss_mask} for the supervision of transient mask,
\begin{equation}
    \mathcal{L}_\textrm{MLP} = \lambda_\textrm{residual} \hat{\mathcal{L}}_\textrm{residual} + \lambda_\textrm{cos} \hat{\mathcal{L}}_\textrm{cos} + \lambda_\textrm{reg} \mathcal{L}_\textrm{reg},
\end{equation}
where $\hat{\mathcal{L}}_\textrm{residual}$ and $\hat{\mathcal{L}}_\textrm{cos}$ denote the loss computed from the residual and the feature similarity, which are calculated using the candidate with the smaller photometric error between the raw rendered image and the affine rendered image.

\section{Experiments}
\label{sec:Experiments}

\begin{table*}[!t] \centering
    \begin{center}
    \captionof{table}{Quantitative results on RobustNeRF dataset~\cite{sabour2023robustnerf}. The best results are highlighted in \textbf{bold}, and the second in \underline{underline}.}
    
\resizebox{\textwidth}{!}{
\begin{tabular}{l|*{3}{c}|*{3}{c}|*{3}{c}|*{3}{c}|*{3}{c}}
    \toprule
    \multirow{2}{*}{Method}
    & \multicolumn{3}{c|}{Android} 
    & \multicolumn{3}{c|}{Crab2}
    & \multicolumn{3}{c|}{Statue}
    & \multicolumn{3}{c|}{Yoda}
    & \multicolumn{3}{c}{Mean}
    \\
    & \multicolumn{1}{c}{PSNR} 
    & \multicolumn{1}{c}{SSIM} 
    & \multicolumn{1}{c|}{LPIPS} 
    & \multicolumn{1}{c}{PSNR} 
    & \multicolumn{1}{c}{SSIM} 
    & \multicolumn{1}{c|}{LPIPS} 
    & \multicolumn{1}{c}{PSNR} 
    & \multicolumn{1}{c}{SSIM} 
    & \multicolumn{1}{c|}{LPIPS} 
    & \multicolumn{1}{c}{PSNR} 
    & \multicolumn{1}{c}{SSIM} 
    & \multicolumn{1}{c|}{LPIPS} 
    & \multicolumn{1}{c}{PSNR} 
    & \multicolumn{1}{c}{SSIM} 
    & \multicolumn{1}{c}{LPIPS} 
    \\
    \midrule
    3DGS~\cite{kerbl20233d}
    & 23.32 & 0.794 & 0.159 
    & 31.76 & 0.925 & 0.172 
    & 20.83 & 0.830 & 0.148 
    & 28.92 & 0.905 & 0.192
    & 26.21 & 0.864 & 0.168
    \\
    SpotLessSplat~\cite{sabour2024spotlesssplats}
    & 24.20 & 0.810 & 0.159 
    & 33.90 & 0.933 & 0.169 
    & 21.97 & 0.821 & 0.163 
    & \underline{34.24} & 0.938 & 0.156 
    & 28.58 & 0.875 & 0.162
    \\
    WildGaussians~\cite{kulhanek2024wildgaussians}
    & \textbf{24.67} & \underline{0.828} & 0.151 
    & 30.52 & 0.909 & 0.213 
    & 22.54 & \underline{0.863} & 0.129 
    & 30.55 & 0.905 & 0.202
    & 27.07 & 0.876 & 0.174
    \\
    Robust3DGaussians~\cite{ungermann2024robust}
    & 24.30 & 0.813 & \underline{0.134}
    & 32.77 & 0.926 & 0.162 
    & 21.93 & 0.837 & 0.135 
    & 30.85 & 0.913 & 0.177 
    & 27.46 & 0.872 & {0.152}
    \\
    DeSplat~\cite{wang2024desplat}
    & 24.26 & 0.815 & 0.150  
    & \underline{34.15} & \underline{0.934} & \textbf{0.153}
    & \textbf{22.80} & 0.844 & \underline{0.125} 
    & \underline{34.24} & \underline{0.939} & \underline{0.153} 
    & \underline{28.89} & \underline{0.883} & \underline{0.146}
    
    \\
    RobustSplat
    & \underline{24.62} & \textbf{0.831} & \textbf{0.125} 
    & \textbf{34.88} & \textbf{0.940} & \underline{0.154} 
    & \textbf{22.80} & \textbf{0.865} & \textbf{0.110} 
    & \textbf{35.14} & \textbf{0.944} & \textbf{0.151} 
    & \textbf{29.36} & \textbf{0.895} & \textbf{0.135}
    \\
    \bottomrule
\end{tabular}
}

    \label{tab:robustnerf}
    \end{center}
    \vspace{-5pt}
    \makebox[0.195\textwidth]{\footnotesize WildGaussians~\cite{kulhanek2024wildgaussians}}
\makebox[0.195\textwidth]{\footnotesize SpotLessSplats~\cite{sabour2024spotlesssplats}}
\makebox[0.195\textwidth]{\footnotesize DeSplat~\cite{wang2024desplat}}
\makebox[0.195\textwidth]{\footnotesize RobustSplat}
\makebox[0.195\textwidth]{\footnotesize GT}
\\
\includegraphics[width=0.195\textwidth]{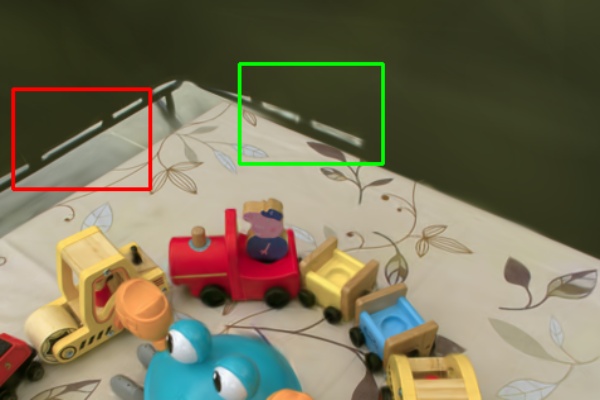}
\includegraphics[width=0.195\textwidth]{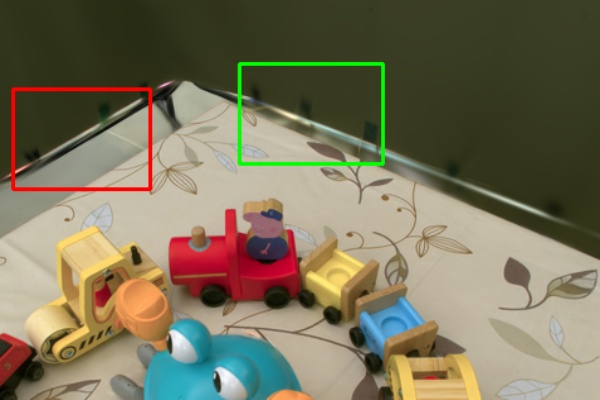}
\includegraphics[width=0.195\textwidth]{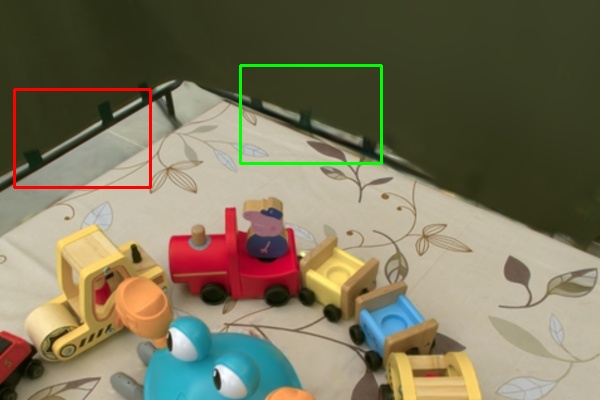}
\includegraphics[width=0.195\textwidth]{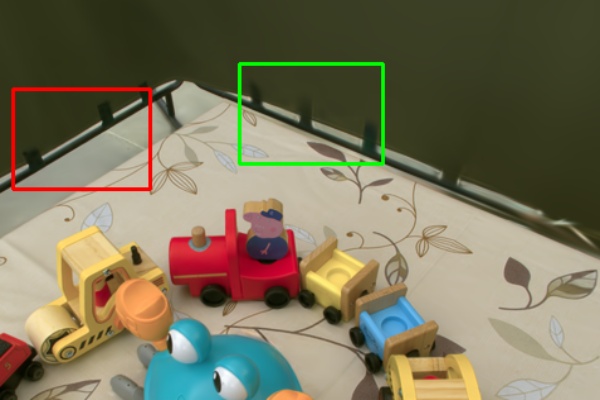}
\includegraphics[width=0.195\textwidth]{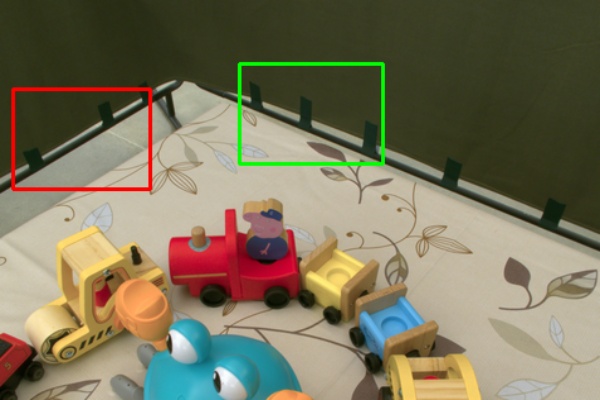}
\\
\includegraphics[width=0.0948\textwidth,height=0.07\textwidth]{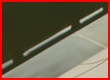}
\includegraphics[width=0.0948\textwidth,height=0.07\textwidth]{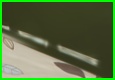}
\includegraphics[width=0.0948\textwidth,height=0.07\textwidth]{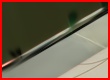}
\includegraphics[width=0.0948\textwidth,height=0.07\textwidth]{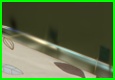}
\includegraphics[width=0.0948\textwidth,height=0.07\textwidth]{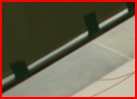}
\includegraphics[width=0.0948\textwidth,height=0.07\textwidth]{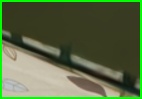}
\includegraphics[width=0.0948\textwidth,height=0.07\textwidth]{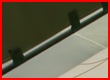}
\includegraphics[width=0.0948\textwidth,height=0.07\textwidth]{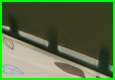}
\includegraphics[width=0.0948\textwidth,height=0.07\textwidth]{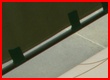}
\includegraphics[width=0.0948\textwidth,height=0.07\textwidth]{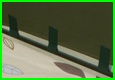}
\\[0.5em]
\includegraphics[width=0.195\textwidth]{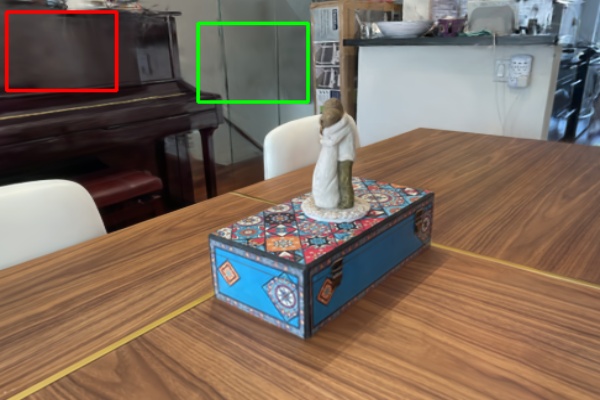}
\includegraphics[width=0.195\textwidth]{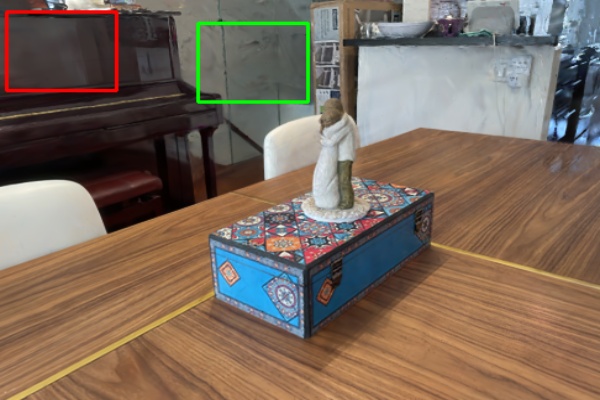}
\includegraphics[width=0.195\textwidth]{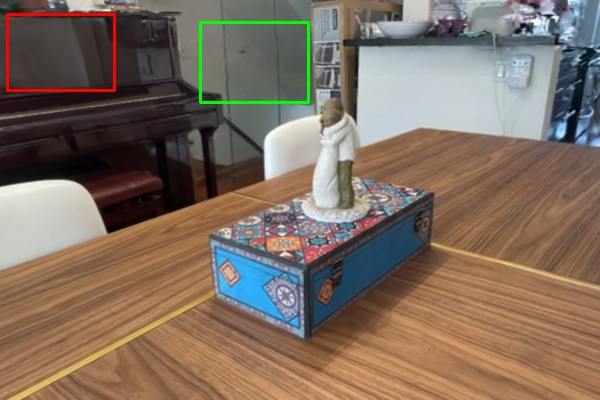}
\includegraphics[width=0.195\textwidth]{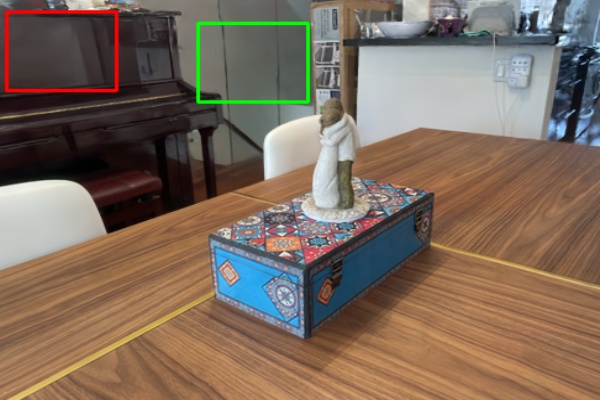}
\includegraphics[width=0.195\textwidth]{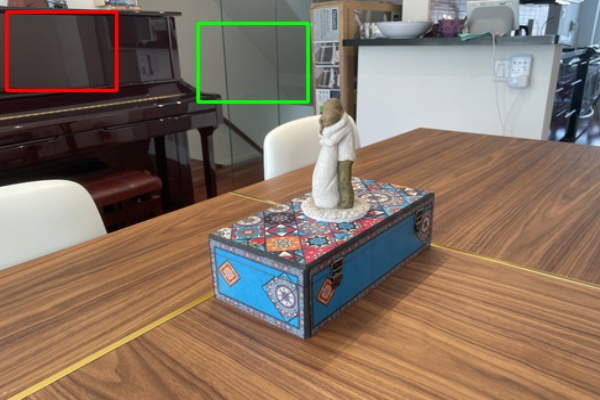}
\\
\includegraphics[width=0.0948\textwidth,height=0.07\textwidth]{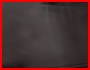}
\includegraphics[width=0.0948\textwidth,height=0.07\textwidth]{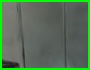}
\includegraphics[width=0.0948\textwidth,height=0.07\textwidth]{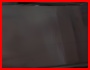}
\includegraphics[width=0.0948\textwidth,height=0.07\textwidth]{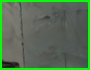}
\includegraphics[width=0.0948\textwidth,height=0.07\textwidth]{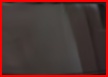}
\includegraphics[width=0.0948\textwidth,height=0.07\textwidth]{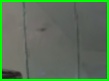}
\includegraphics[width=0.0948\textwidth,height=0.07\textwidth]{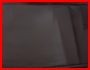}
\includegraphics[width=0.0948\textwidth,height=0.07\textwidth]{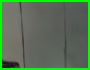}
\includegraphics[width=0.0948\textwidth,height=0.07\textwidth]{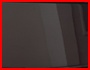}
\includegraphics[width=0.0948\textwidth,height=0.07\textwidth]{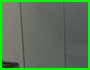}
\\

    \captionof{figure}{Qualitative comparison on \emph{Crab2} and \emph{Statue} from RobustNeRF dataset~\cite{sabour2023robustnerf}.}
    \label{fig:qual_robustnerf}
    \vspace{-1em}
\end{table*}

\paragraph{Datasets} 
We evaluate our RobustSplat on two challenging benchmark datasets: \emph{NeRF On-the-go}~\cite{ren2024nerf} and \emph{RobustNeRF}~\cite{sabour2023robustnerf}. In addition, we further validate the accuracy in handling transients on clean datasets, including \emph{RobustNeRF clean}~\cite{sabour2023robustnerf} and \emph{Mip-NeRF 360}~\cite{barron2022mip}.

To validate the robustness under illumination variations, we evaluate our RobustSplat++ on two challenging benchmark datasets: \emph{NeRF-OSR}~\cite{rudnev2022nerfosr} and \emph{PhotoTourism}~\cite{2006phototourism}.

\paragraph{Implementation Details} Our codebase follows the official Gaussian Splatting (3DGS). During training, we adopt the same learning rate settings as 3DGS and set the total training iterations to 30K. The $\textrm{MLP}_\textrm{mask}$ consists of two linear layers, optimized with the Adam optimizer with a learning rate of 0.001. 
Fixed parameters are used for all experiments. 
The delayed iteration start is set to 10K, and the weights for the $\textrm{MLP}_\textrm{mask}$ supervision terms $\lambda_\textrm{residual}$, $\lambda_\textrm{cos}$, $\lambda_\textrm{reg}$ are set to 0.5, 0.5, 2.0, respectively.
The mask regularization coefficient $\beta_\textrm{reg}$ is set to 2000. {The sensitivity of the hyper-parameters is analyzed in \sref{sec:ablation_transient}.} The features used by the $\textrm{MLP}_\textrm{mask}$ are extracted from DINOv2, with pre-trained weights from \emph{ViT-S/14 distilled}.

In the mask bootstrapping, the lowest spatial resolution features are extracted from images of size $(224\times224)$, while the highest spatial resolution features are derived from size $(504\times504)$. 

Following existing methods~\cite{sabour2024spotlesssplats}, we apply a downsampling factor of 8 on the NeRF On-the-go and RobustNeRF datasets (factor 4 for specific scenarios, e.g., arcdetriomphe and patio). 
For the Mip-NeRF 360 dataset, we adopt a downsampling factor of 2 for the indoor scenes and a factor of 4 for the outdoor scenes.
Low-resolution residuals are further downsampled by an additional factor of 4 based on this configuration.

In the appearance modeling, the $\textrm{MLP}_\textrm{affine}$ consists of three linear layers and is optimized using the Adam optimizer with a learning rate of 0.0005. The 2D image embeddings $f_\textrm{img}$ have a dimension of 48, are initialized from a normal distribution with a standard deviation of 0.01. Its learning rate is set to 0.001 during training and 0.01 for test-time optimization. The 3D Gaussian embeddings $f_\textrm{gs}$ have a dimension of 30, and are initialized by applying Fourier-based positional encoding of the initial SfM points following~\cite{mildenhall2020nerf,kulhanek2024wildgaussians}, with a learning rate of 0.005.

\subsection{Evaluation for \textbf{Transient-free} 3DGS}
\label{sub:Results I}

\paragraph{Baselines} We evaluated our RobustSplat against {multiple baselines, including the vanilla 3D Gaussian Splatting~\cite{kerbl20233d} which we built upon, and recent robust methods including SpotLessSplats~\cite{sabour2024spotlesssplats}, WildGaussians~\cite{kulhanek2024wildgaussians}, Robust3DGaussians~\cite{ungermann2024robust} and DeSplat~\cite{wang2024desplat}.
To ensure a fair comparison, we utilized the publicly available implementations of these methods and conducted evaluations using the same camera matrices across all experiments. 
We assessed performance through both visual comparisons and quantitative metrics, employing PSNR, SSIM, and LPIPS to measure reconstruction quality.
}

\begin{table}[t] \centering
    \begin{center}
    \captionof{table}{Evaluation on Mip-NeRF 360 dataset~\cite{barron2022mip} and RobustNeRF Clean dataset~\cite{sabour2023robustnerf}.}
    \label{table:clean_comparison}
    \resizebox{0.48\textwidth}{!}{
    \begin{tabular}{l|ccc|ccc}
    \toprule
    \multirow{2}{*}{Method} 
    & \multicolumn{3}{c|}{MipNeRF 360} 
    & \multicolumn{3}{c}{RobustNeRF Clean} 
\\
    & PSNR & SSIM & LPIPS 
    & PSNR & SSIM & LPIPS
\\
    \midrule
    3DGS~\cite{kerbl20233d}
    & \textbf{27.78} & \textbf{0.826} & \textbf{0.202} 
    & \textbf{32.14} & \textbf{0.925} & \textbf{0.107} 
\\
    SpotLessSplats~\cite{sabour2024spotlesssplats}
    & 25.52 & 0.727 & 0.311 
    & 29.69 & 0.901 & 0.140 
\\
    RobustSplat 
    & 27.30 & 0.798 & 0.251 
    & 32.12 & 0.922 & 0.113 
\\
    \bottomrule
    \end{tabular}
}

    \end{center}
    \vspace{-2pt}

\makebox[0.24\textwidth]{\footnotesize SpotLessSplats}
\makebox[0.24\textwidth]{\footnotesize RobustSplat}
\\
\includegraphics[width=0.135\textwidth]{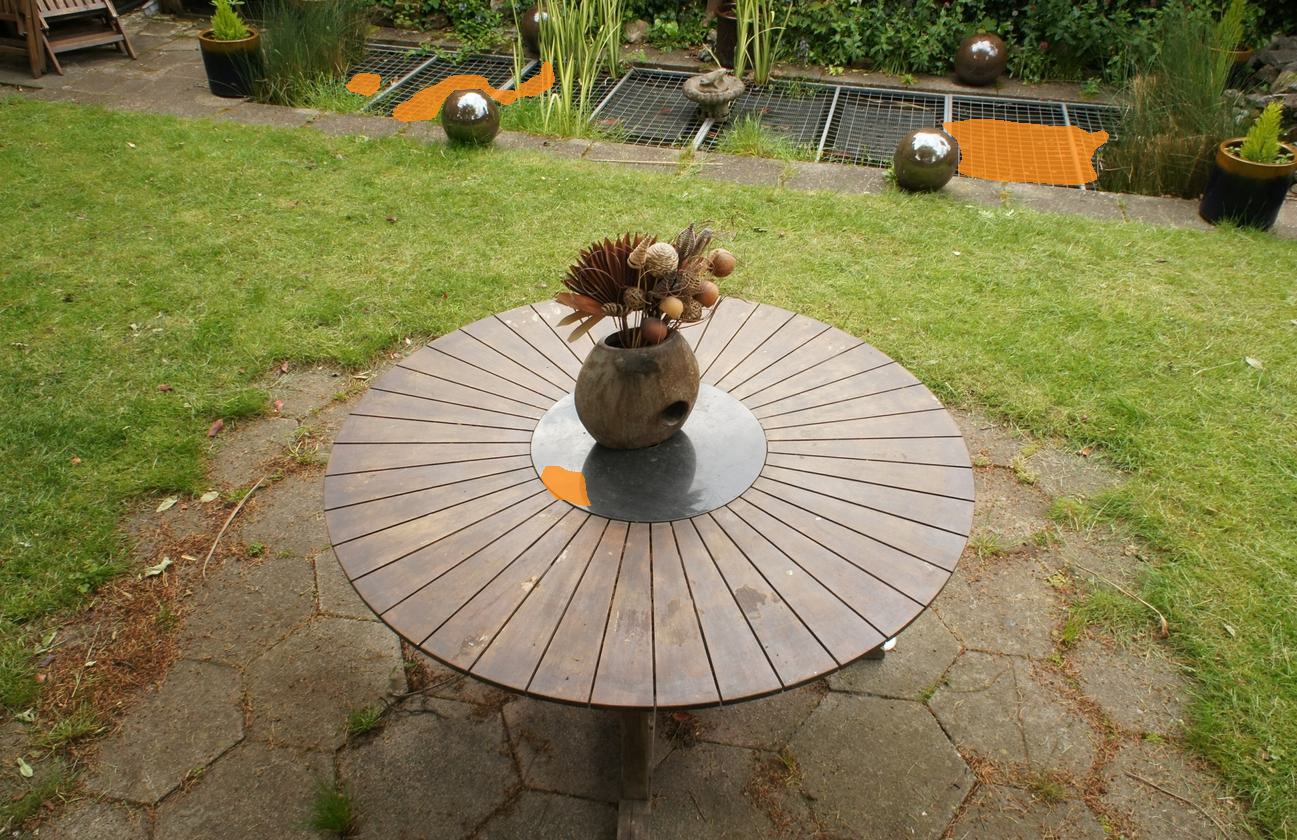}
\includegraphics[width=0.102\textwidth]{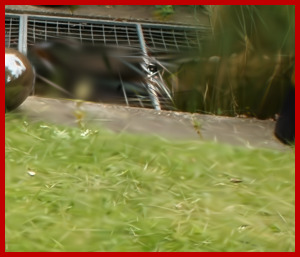}
\includegraphics[width=0.135\textwidth]{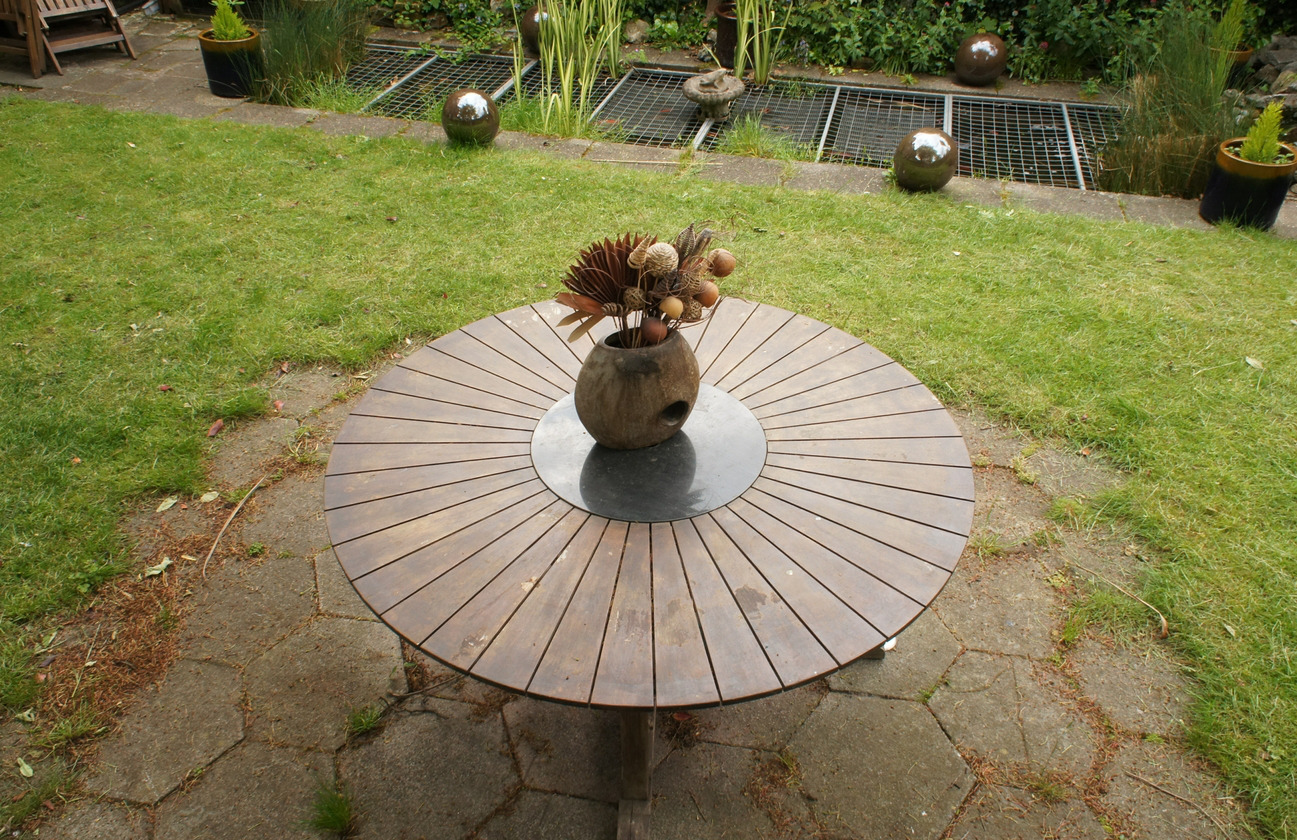}
\includegraphics[width=0.102\textwidth]{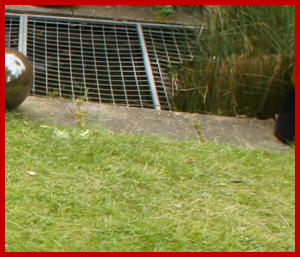}
\\
\makebox[0.24\textwidth]{\footnotesize (a) \emph{Garden} from MipNeRF 360 Outdoor}
\\[0.3em]
\includegraphics[width=0.135\textwidth]{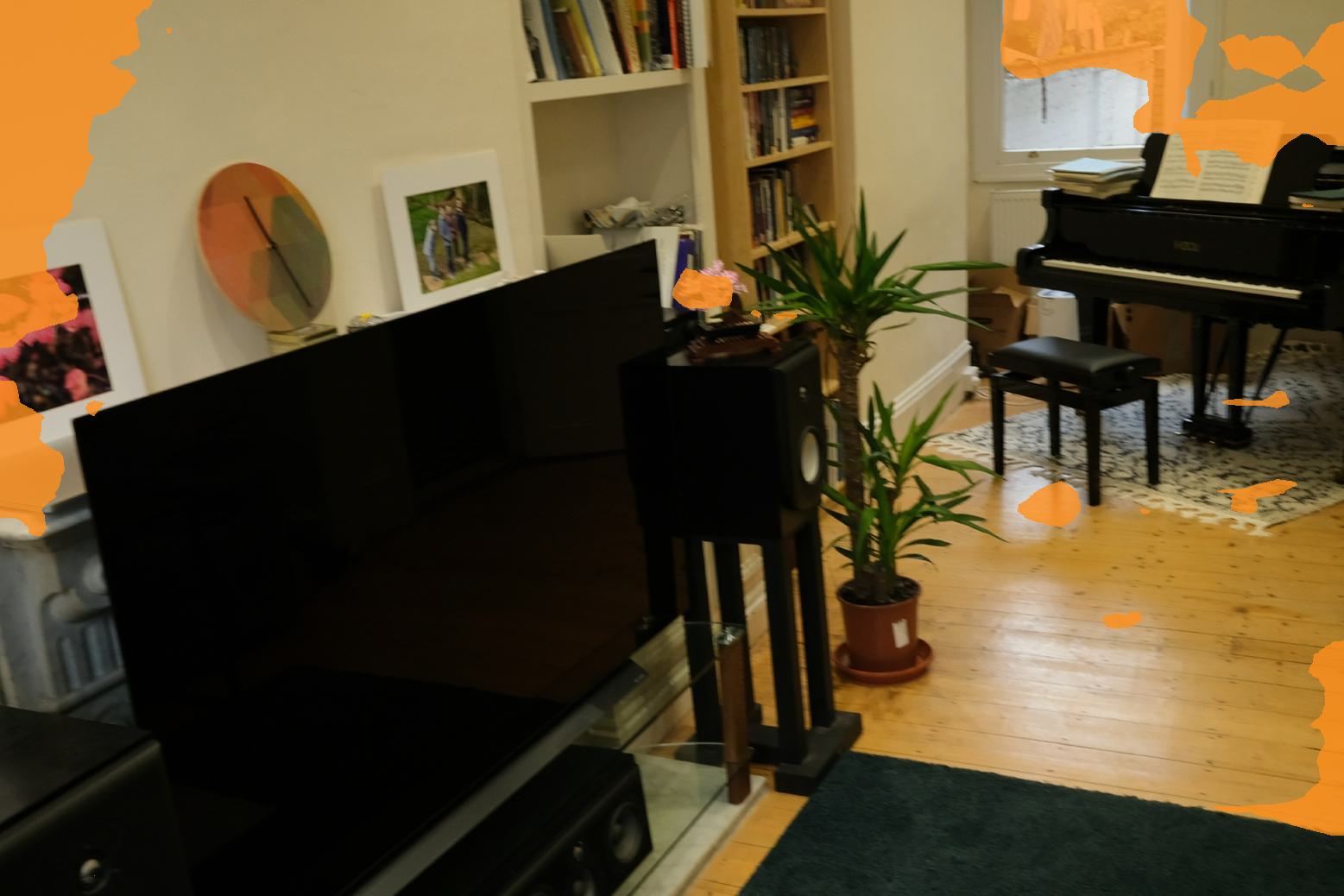}
\includegraphics[width=0.102\textwidth]{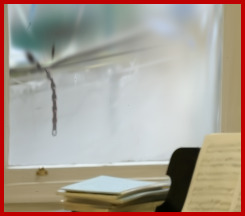}
\includegraphics[width=0.135\textwidth]{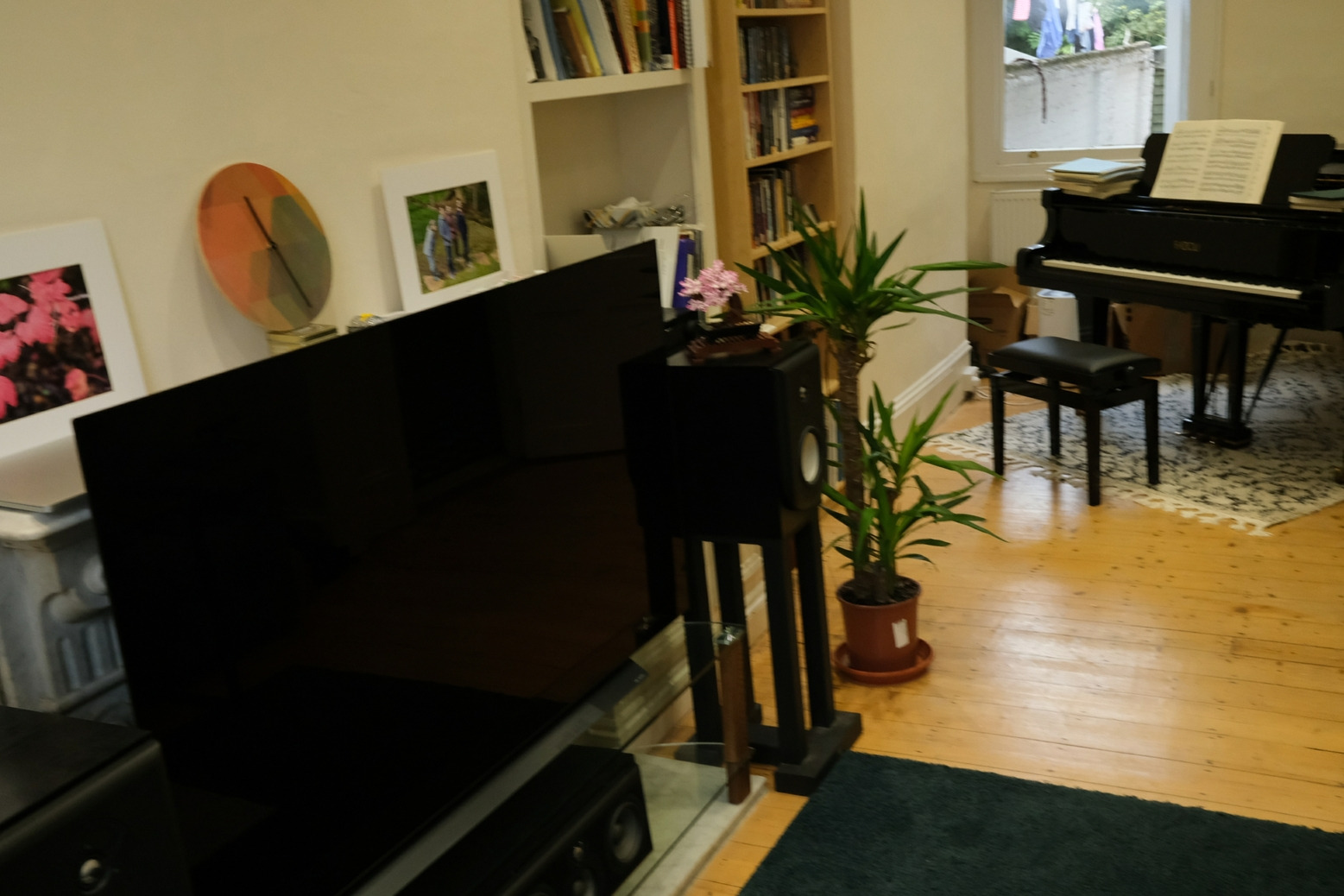}
\includegraphics[width=0.102\textwidth]{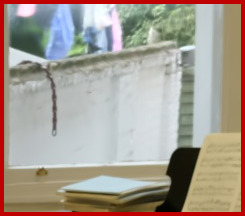}
\\
\makebox[0.24\textwidth]{\footnotesize (b) \emph{Room} from MipNeRF 360 Indoor}
\\[0.3em]
\includegraphics[width=0.135\textwidth]{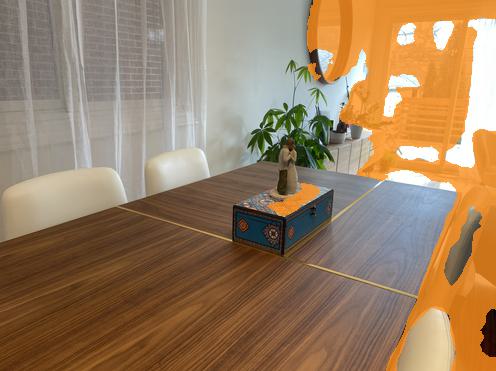}
\includegraphics[width=0.102\textwidth]{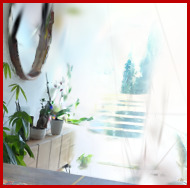}
\includegraphics[width=0.135\textwidth]{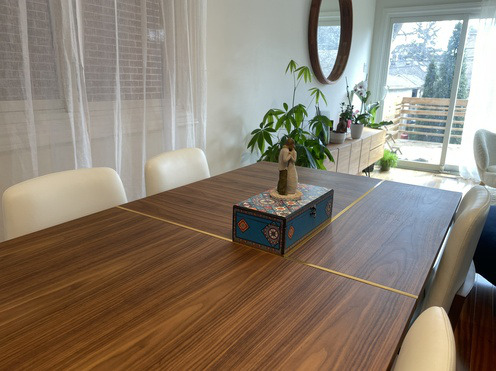}
\includegraphics[width=0.102\textwidth]{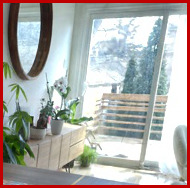}
\\
\makebox[0.24\textwidth]{\footnotesize (c) \emph{Statue} from RobustNeRF clean}

    \captionof{figure}{Qualitative comparison of predicted masks and zoom-in rendered images on \emph{garden}, \emph{room} from Mip-NeRF 360 dataset~\cite{barron2022mip} and \emph{statue} from RobustNeRF Clean dataset~\cite{sabour2023robustnerf}.} 
    \label{fig:clean}
    \vspace{-1em}
\end{table}

\paragraph{\textbf{\emph{NeRF On-the-go}} Dataset}
We first evaluate our method on NeRF On-the-go dataset. We can see from \Tref{table:onthego} that our method achieves the best results across all six scenes on the PSNR, SSIM, and LPIPS metrics, clearly demonstrating the effectiveness of our method.

\Fref{fig:qual_onthego} shows the qualitative comparison, in which baseline approaches exhibit noticeable artifacts. Thanks to the proposed \delayedGS and \coarsetofine design, our method successfully eliminates these artifacts and achieves superior detail (e.g., the windows in \emph{Patio-high}, as well as the building in \emph{Fountain}).

\newcommand{\psnr}{\text{PSNR}}
\newcommand{\ssim}{\text{SSIM}}
\newcommand{\lpips}{\text{LPIPS}}

\paragraph{\textbf{\emph{RobustNeRF}} Dataset}
\label{sub:Results II}
To further validate the effectiveness of our method, we conduct comparisons with baseline methods on the RobustNeRF dataset, with quantitative results shown in \Tref{tab:robustnerf}. 
Our method achieves the best performance on the average metric.
Although our method performs slightly worse in PSNR metric on the \emph{Android} scene, it remains competitive with the state-of-the-art methods. 
In the remaining three scenes of the RobustNeRF dataset, our approach significantly outperforms existing methods. 
The qualitative comparisons are presented in \fref{fig:qual_robustnerf}, which shows that our method achieves transient-free reconstruction with sharp details.

\paragraph{\textbf{\emph{Clean}} Data}
\label{sub:Results III}
{To evaluate the accuracy of mask estimation and the scene representation capability of our method, we further conduct clean data comparisons with baseline methods on the RobustNeRF clean dataset and the MipNeRF 360 dataset. The quantitative results in \Tref{table:clean_comparison} show that our method achieves results comparable to 3DGS, and significantly outperforms SpotLessSplats. The visual comparison of estimated mask and rendering are presented in \fref{fig:clean}, which shows that our method has fewer errors in mask prediction and better quality in detail rendering compared to SpotLessSplats.}

\subsection{Ablation Study for \textbf{Transient-free} 3DGS}
\label{sec:ablation_transient}
To evaluate the effectiveness of each component of our method, we built upon the 3DGS~\cite{kerbl20233d} and added different components to analyze the model performance.

\paragraph{Effects of Delayed Gaussian Growth} \Tref{table:ablation_method} shows that compared with the full model, the model without \delayedGS (``3DGS+Mask+MB'') suffers from a noticeable decrease in all average metrics, which reiterates the effectiveness of the \delayedGS strategy in preventing the 3DGS from fitting transient regions during the early optimization phase.

\paragraph{Effects of Mask Bootstrapping} 
\Tref{table:ablation_method} shows that removing the proposed \coarsetofine (``3DGS+Mask+DG'') leads to a decrease in overall performance. This drop is particularly evident in the Mountain scene, an unbounded environment with a large proportion of sky regions and sparsely initialized points, which leading to overly smooth reconstructions during early optimization. Our mask bootstrapping provides more robust supervision, leading to more accurate reconstructions.

\paragraph{Effects of Mask Regularization} 
{The mask regularization is useful for stabilizing early-stage training due to} unconverged reconstruction and significant photometric errors in early training stages. \Tref{table:maskreg} shows that removing the proposed mask regularization leads to a decrease in overall performance, which indicates that our simple yet effective warm-up strategy is helpful for early-stage mask learning.

\paragraph{Sensitivity Analysis}
{To verify the robustness of our proposed method to variations in key hyperparameters, we conduct a} sensitivity analysis on a core hyperparameter that directly affects the mask prediction. The hyperparameter $\beta_\textrm{reg}$ controls the decay rate of the mask learning regularization
term $\mathcal{L}_\textrm{reg}$. We analyze the effect of different values of $\beta_\textrm{reg}$, as shown in ~\fref{fig:beta_reg_sensitivity}(a). As $\beta_\textrm{reg}$ increases, the decay rate of the curve slows down, which indicating that the warm-up time of the mask prediction is extended. \fref{fig:beta_reg_sensitivity}(b) shows the overall performance of our method across different values of $\beta_\textrm{reg}$, with results reported in terms of PSNR and SSIM. The results indicate that our method maintains stable performance across the tested $\beta_\textrm{reg}$ values, and the best performance is achieved when $\beta_\textrm{reg}$ is set to 2000, where regularization is disabled around 6K iterations.

\begin{table}[t]
    \begin{center}
    \caption{Ablation of each component in our method on NeRF On-the-go datasets~\cite{ren2024nerf}. ``3DGS+Mask'' is the model that integrate the transient mask estimation with 3DGS. We denote \emph{Mask Bootstrapping} as ``MB'', and \emph{\DelayedGS} as ``DG''. ``{RobustSplat}'' indicates ``3DGS+Mask+MB+DG''.}
    \label{table:ablation_method}
    \resizebox{0.48\textwidth}{!}{
    \begin{tabular}{l|cc|cc|cc}
    \toprule
    \multirow{2}{*}{Method} 
    & \multicolumn{2}{c|}{Low Occlusion}  
    & \multicolumn{2}{c|}{Medium Occlusion} 
    & \multicolumn{2}{c}{High Occlusion} \\
    & PSNR & SSIM 
    & PSNR & SSIM 
    & PSNR & SSIM \\
    \midrule
    3DGS~\cite{kerbl20233d}
    & 19.65 & 0.689
    & 19.35 & 0.774
    & 17.79 & 0.687\\
    \phantom{00}+ Mask
    & 20.28 & 0.696 
    & 23.14 & 0.849 
    & 23.47 & 0.868 \\
    \phantom{0000}+ DG
    & 20.92 & 0.711 
    & 23.75 & 0.862 
    & 24.18 & \textbf{0.872} \\
    \phantom{0000}+ MB
    & 20.81 & 0.703 
    & 23.20 & 0.851 
    & 23.68 & 0.863 \\
    RobustSplat 
    & \textbf{21.08} & \textbf{0.719}
    & \textbf{24.03} & \textbf{0.862}
    & \textbf{24.54} & \textbf{0.872}\\
    \bottomrule
    \end{tabular}}

    \end{center}
\end{table}

\begin{table}[t] \centering
    \caption{Effects of Mask Regularization (denote as ``MR'').}
    \vspace{-0.2em}
    \label{table:maskreg}
    \resizebox{0.48\textwidth}{!}{
    \begin{tabular}{l|cc|cc|cc}
    \toprule
    \multirow{2}{*}{Method} 
    & \multicolumn{2}{c|}{Low Occlusion}  
    & \multicolumn{2}{c|}{Medium Occlusion} 
    & \multicolumn{2}{c}{High Occlusion} \\
     & PSNR & SSIM 
     & PSNR & SSIM 
     & PSNR & SSIM \\
    \midrule
    RobustSplat w/o MR 
    & 20.98 & 0.715 
    & 23.90 & 0.858 
    & 24.16 & \textbf{0.872} \\
    RobustSplat 
    & \textbf{21.08} & \textbf{0.719} 
    & \textbf{24.03} & \textbf{0.862} 
    & \textbf{24.54} & \textbf{0.872} \\
    \bottomrule
    \end{tabular}
}

    \vspace{1em}
    \includegraphics[width=0.48\textwidth]{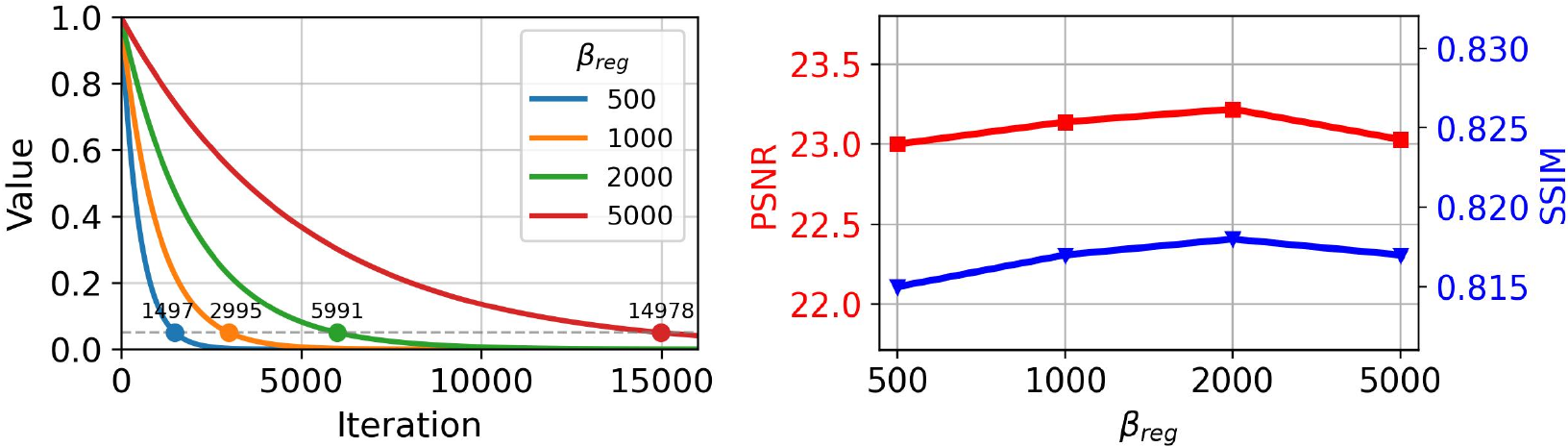}
    \\
    \vspace{-0.4em}
    \makebox[0.245\textwidth]{\scriptsize (a) Curve visualization of $\textrm{exp}{(-i/\beta_\textrm{reg})}$}
    \makebox[0.225\textwidth]{\scriptsize (b) Average metrics} 
    \captionof{figure}{Sensitivity analysis of $\beta_\textrm{reg}$ on NeRF On-the-go dataset~\cite{ren2024nerf}.} 
    \label{fig:beta_reg_sensitivity}
\end{table}

\begin{table*}[!t] \centering
    \begin{center}
    \captionof{table}{Quantitative results on NeRF-OSR dataset~\cite{rudnev2022nerfosr} and PhotoTourism dataset~\cite{2006phototourism}. The best results are highlighted in \textbf{bold}, and the second in \underline{underline}.}
    \resizebox{\textwidth}{!}{
\begin{tabular}{l|*{3}{c}|*{3}{c}|*{3}{c}|*{3}{c}|*{3}{c}|*{3}{c}}
    \toprule
    \multirow{3}{*}{Method}
    & \multicolumn{9}{c|}{NeRF-OSR} 
    & \multicolumn{9}{c}{PhotoTourism} \\
    & \multicolumn{3}{c}{Site 1 (lk2)} 
    & \multicolumn{3}{c}{Site 2 (st)}
    & \multicolumn{3}{c|}{Site 3 (lwp)}
    & \multicolumn{3}{c}{Brandenburg Gate}
    & \multicolumn{3}{c}{Sacre Coeur}
    & \multicolumn{3}{c}{Trevi Fountain}
    \\
    & PSNR & SSIM & LPIPS 
    & PSNR & SSIM & LPIPS
    & PSNR & SSIM & LPIPS
    & PSNR & SSIM & LPIPS
    & PSNR & SSIM & LPIPS
    & PSNR & SSIM & LPIPS
    \\
    \midrule
    3DGS~\cite{kerbl20233d}
    & 15.06 & 0.636 & 0.363  
    & 12.87 & 0.611 & 0.367  
    & 11.58 & 0.583 & \underline{0.377}
    & 20.56 & 0.890 & 0.173
    & 17.17 & 0.831 & 0.209    
    & 16.43 & 0.646 & 0.338 
    \\
    3DGS-E~\cite{kerbl20233d}
    & 16.47 & 0.669 & \underline{0.350} 
    & 13.92 & 0.626 & \underline{0.365}
    & 12.86 & 0.602 & 0.395  
    & 26.32 & 0.920 & 0.150
    & 19.23 & 0.851 & 0.189
    & 17.62 & 0.646 & 0.349 
    \\
    NexusSplats~\cite{ungermann2024robust}
    & \underline{18.85} & 0.658 & 0.383  
    & \underline{16.13} & \underline{0.648} & 0.401  
    & \underline{14.87} & \underline{0.629} & 0.403  
    & 26.40 & 0.920 & 0.149 
    & 21.60 & 0.854 & 0.178
    & \textbf{23.80} & 0.763 & 0.239 
    \\
    DeSplat~\cite{wang2024desplat}
    & 17.42 & 0.664 & 0.399  
    & 15.53 & 0.640 & 0.372  
    & 13.87 & 0.606 & 0.387  
    & 26.58 & 0.916 & 0.139       
    & 21.74 & 0.868 & \underline{0.169}  
    & 21.48 & 0.758 & 0.251 
    \\
    WildGaussians~\cite{kulhanek2024wildgaussians}
    & 17.91 & \underline{0.671} & 0.428  
    & 15.16 & 0.643 & 0.408  
    & 14.56 & 0.620 & 0.396  
    & \underline{27.29} & \underline{0.926} & \underline{0.133}
    & \underline{22.71} & \underline{0.860} & 0.173
    & \underline{23.70} & \underline{0.766} & \underline{0.229}
    \\
    RobustSplat++
    & \textbf{19.13} & \textbf{0.707} & \textbf{0.320}  
    & \textbf{16.54} & \textbf{0.667} & \textbf{0.354}  
    & \textbf{14.88} & \textbf{0.648} & \textbf{0.366}  
    & \textbf{27.63} & \textbf{0.927} & \textbf{0.084}
    & \textbf{22.97} & \textbf{0.870} & \textbf{0.107}
    & 23.28          & \textbf{0.772} & \textbf{0.177}
    \\
    \bottomrule
\end{tabular}
}

    \label{tab:osrpt}
    \end{center}
    \makebox[0.195\textwidth]{\footnotesize 3DGS-E~\cite{kerbl20233d}}
\makebox[0.195\textwidth]{\footnotesize WildGaussians~\cite{kulhanek2024wildgaussians}}
\makebox[0.195\textwidth]{\footnotesize DeSplat~\cite{wang2024desplat}}
\makebox[0.195\textwidth]{\footnotesize RobustSplat++}
\makebox[0.195\textwidth]{\footnotesize GT}
\\
\includegraphics[width=0.195\textwidth]{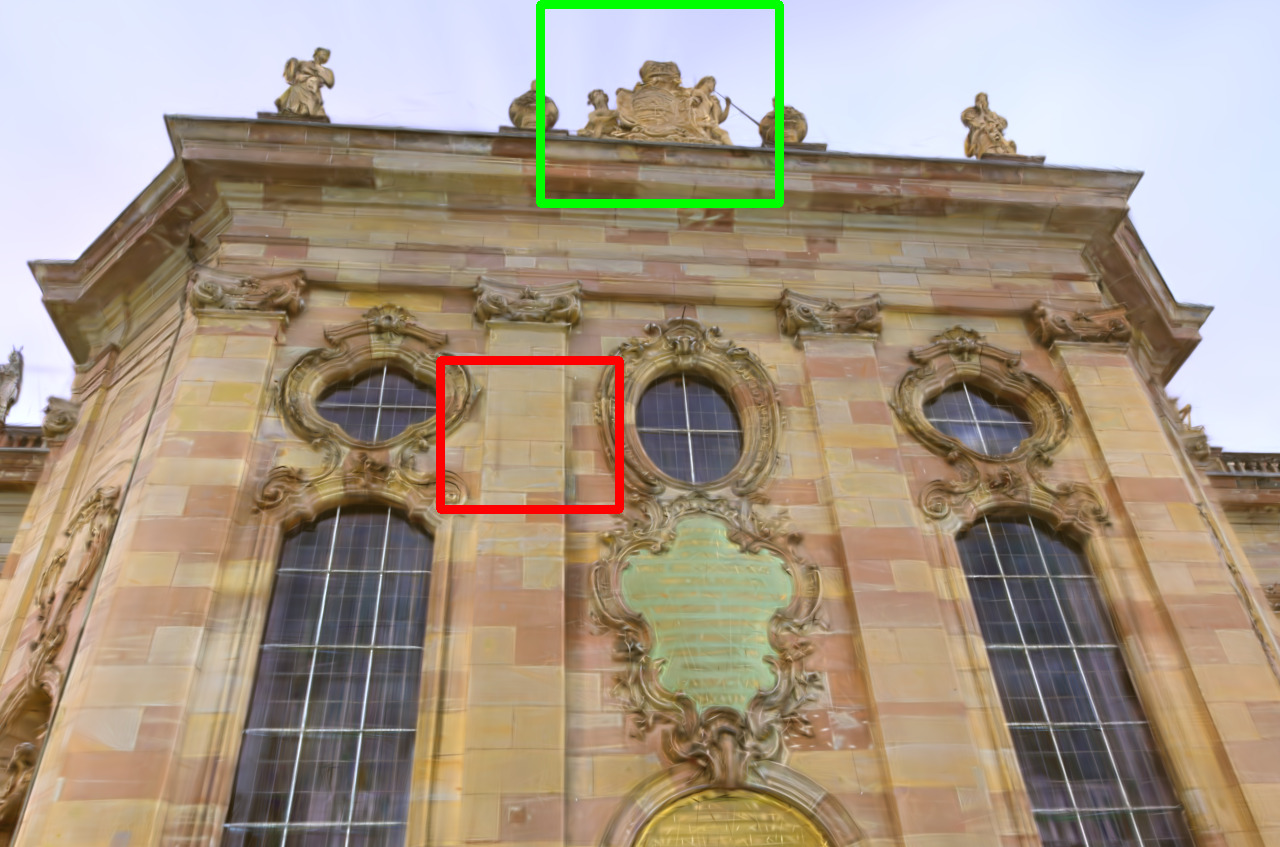}
\includegraphics[width=0.195\textwidth]{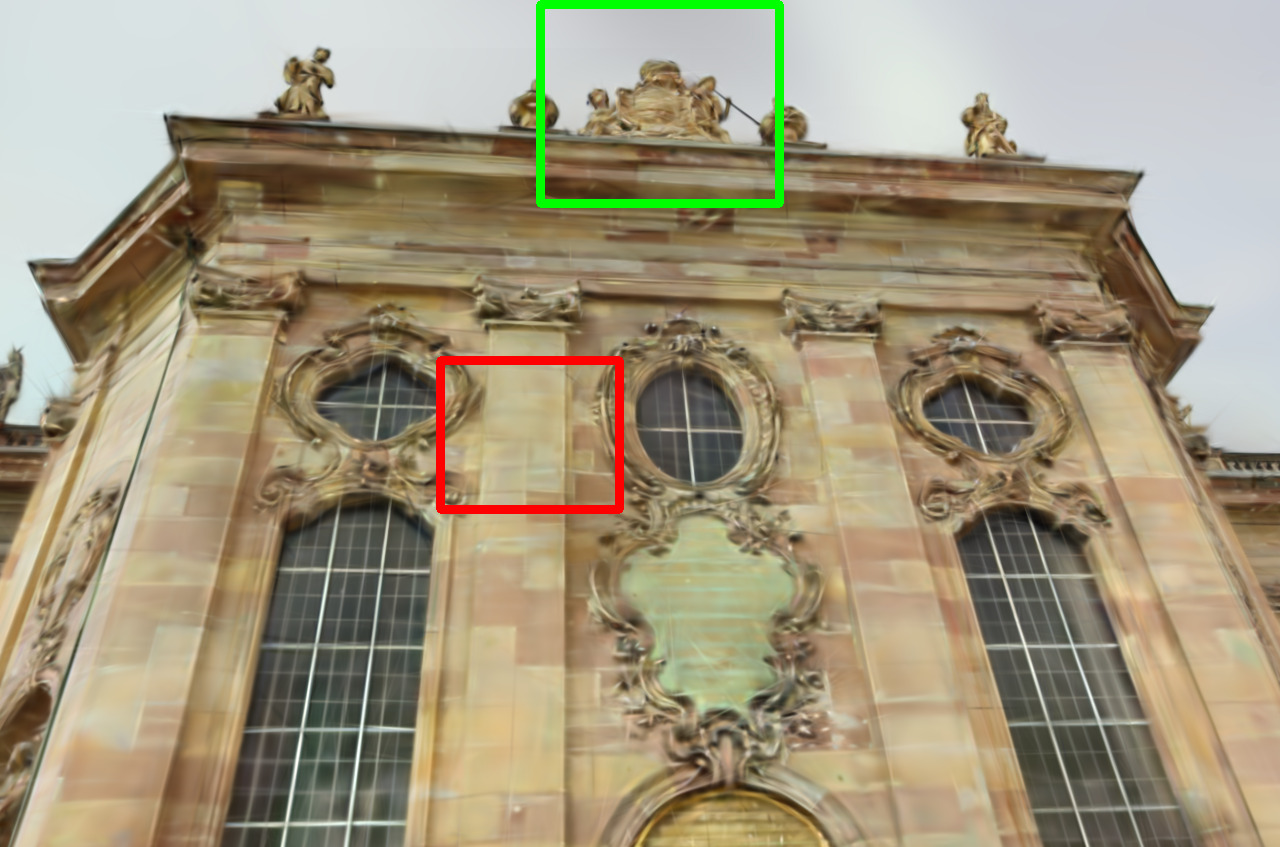}
\includegraphics[width=0.195\textwidth]{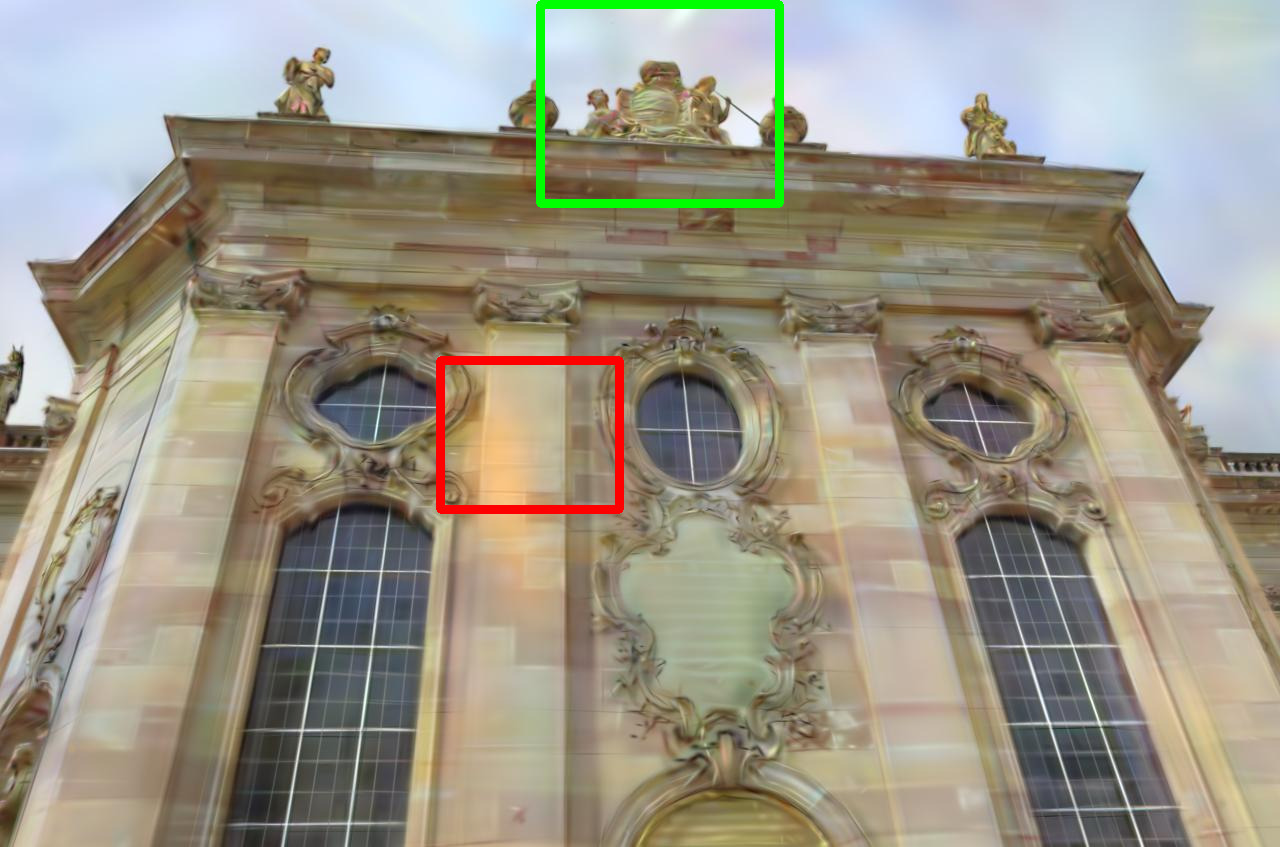}
\includegraphics[width=0.195\textwidth]{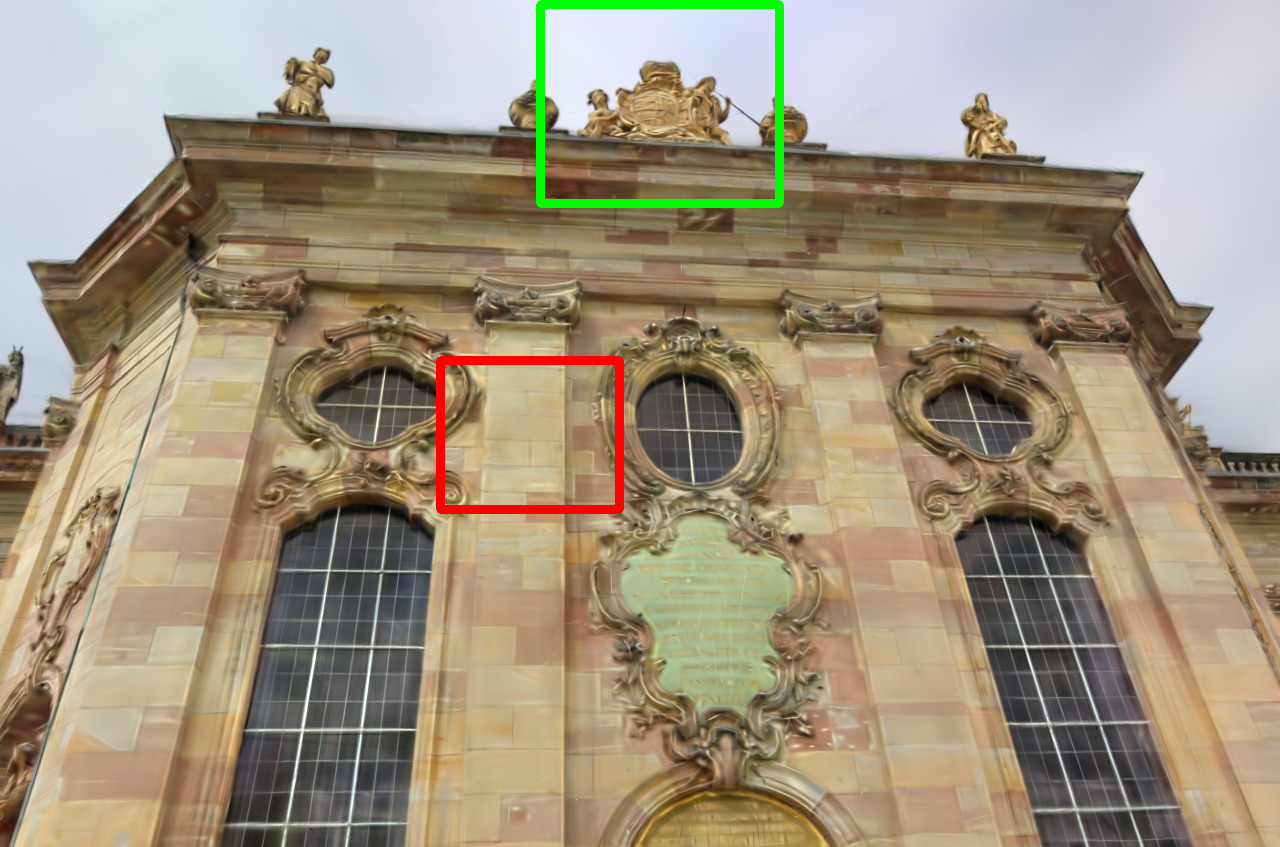}
\includegraphics[width=0.195\textwidth]{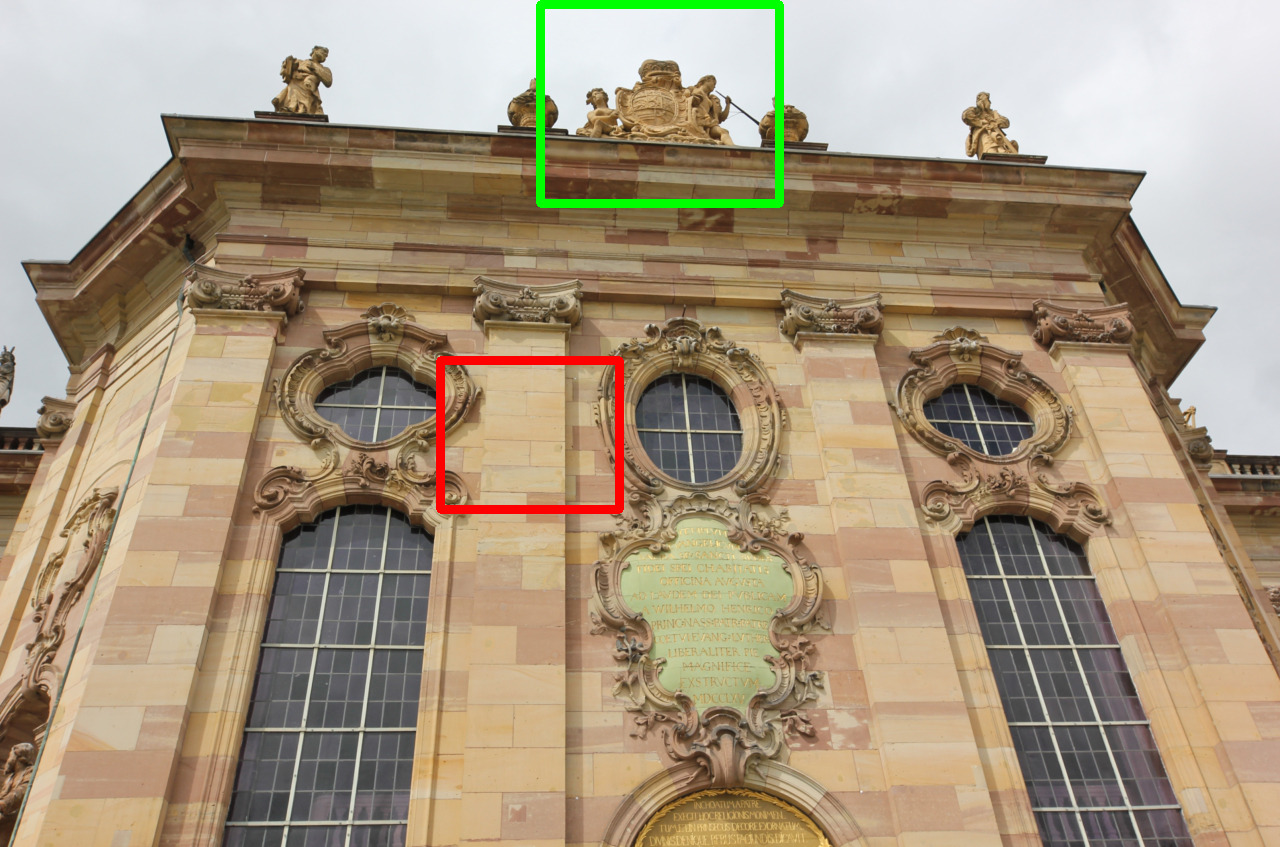}
\\
\includegraphics[width=0.09478\textwidth]{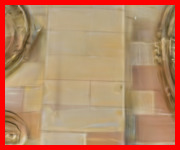}
\includegraphics[width=0.09478\textwidth]{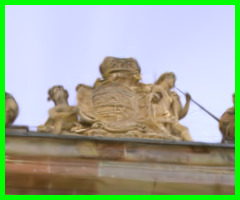}
\includegraphics[width=0.09478\textwidth]{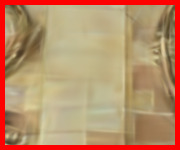}
\includegraphics[width=0.09478\textwidth]{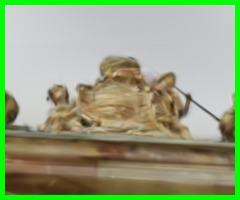}
\includegraphics[width=0.09478\textwidth]{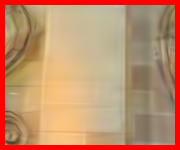}
\includegraphics[width=0.09478\textwidth]{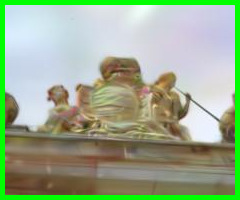}
\includegraphics[width=0.09478\textwidth]{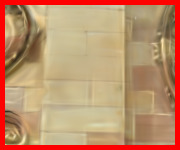}
\includegraphics[width=0.09478\textwidth]{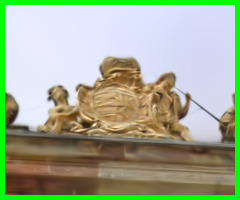}
\includegraphics[width=0.09478\textwidth]{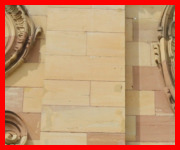}
\includegraphics[width=0.09478\textwidth]{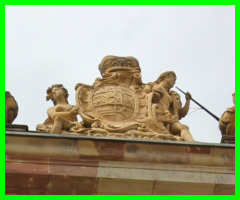}
\\[0.5em]
\includegraphics[width=0.195\textwidth]{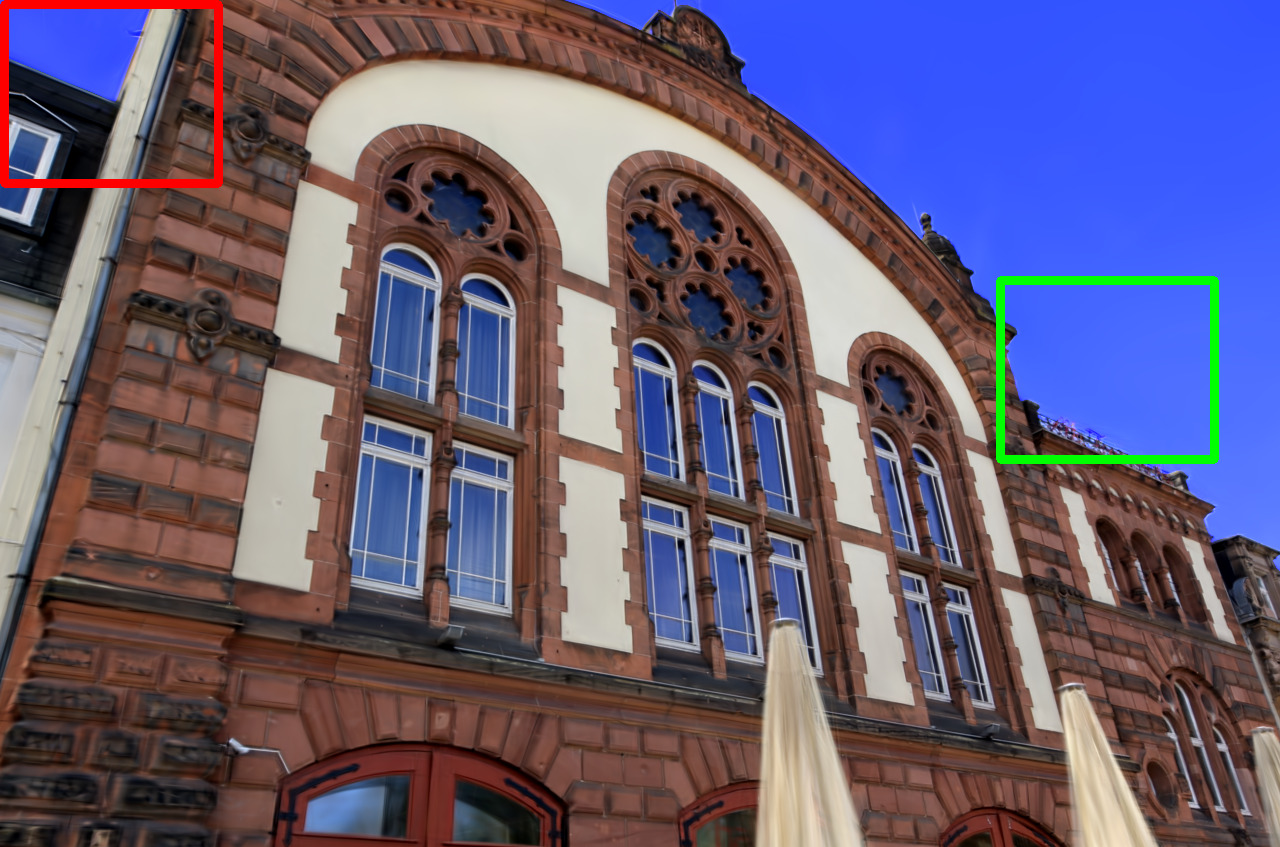}
\includegraphics[width=0.195\textwidth]{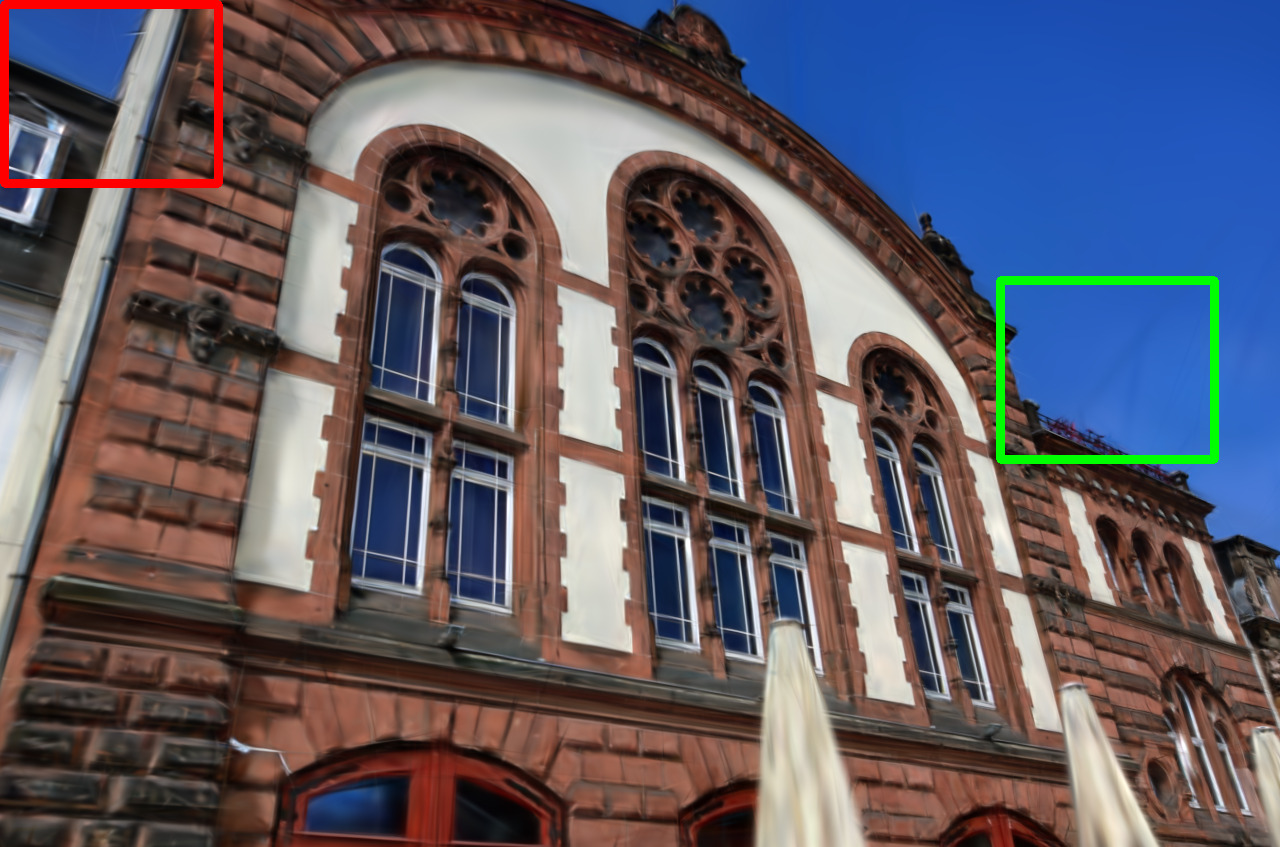}
\includegraphics[width=0.195\textwidth]{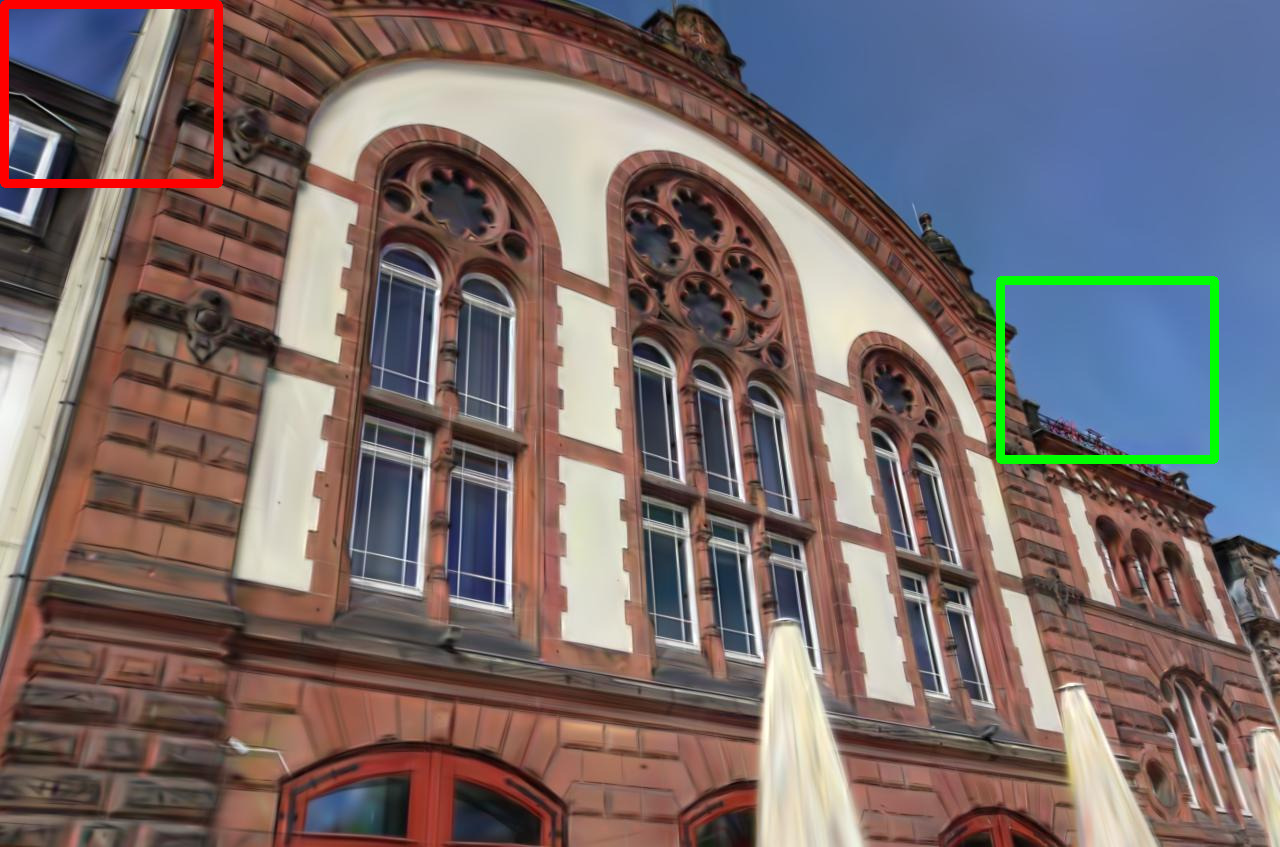}
\includegraphics[width=0.195\textwidth]{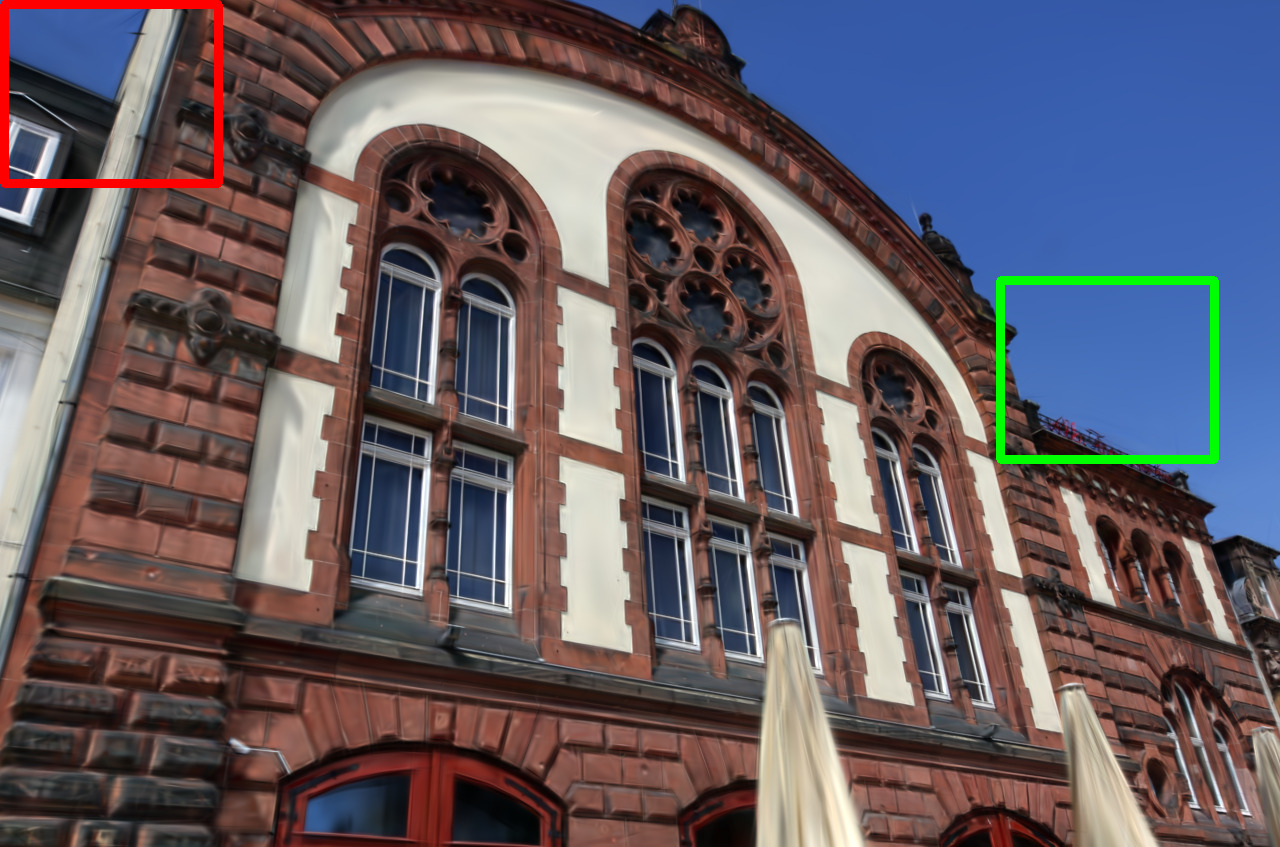}
\includegraphics[width=0.195\textwidth]{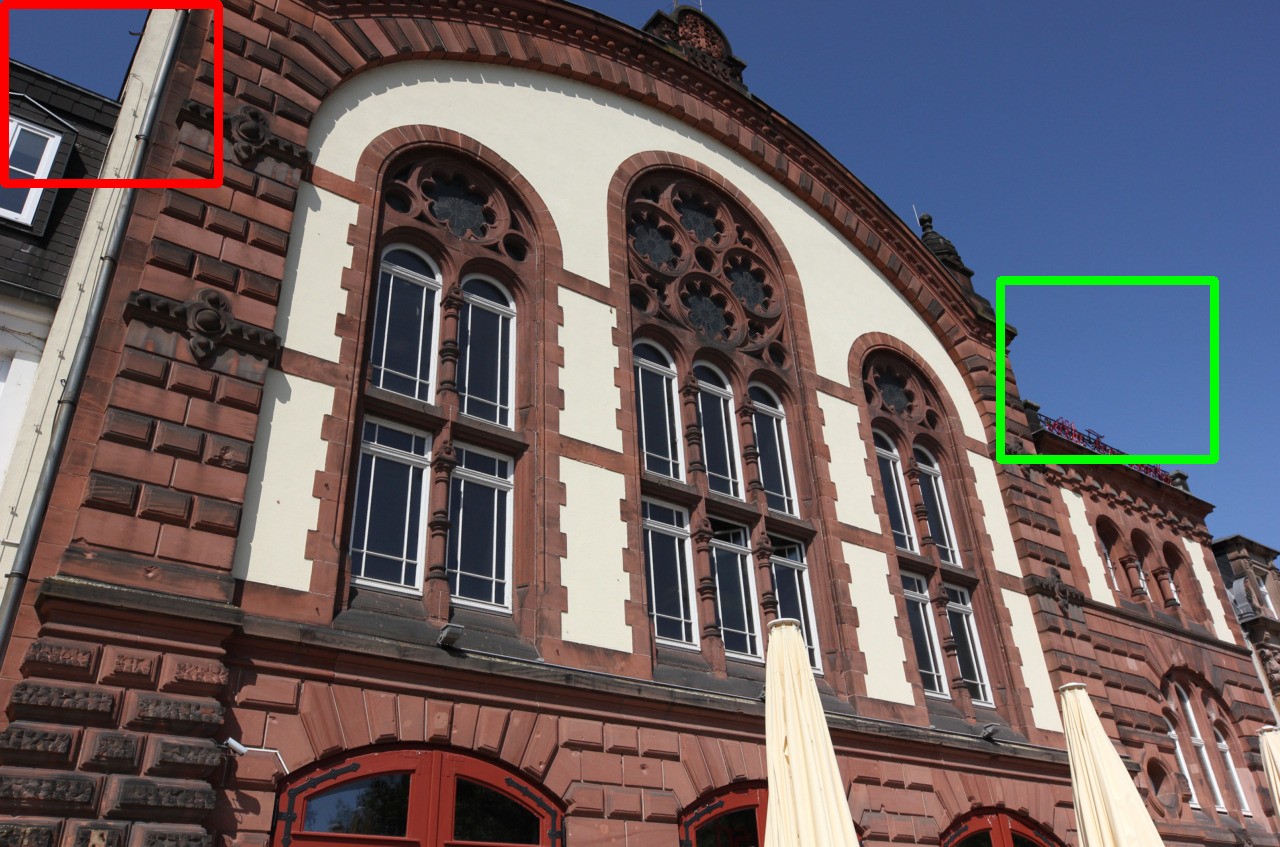}
\\
\includegraphics[width=0.09478\textwidth]{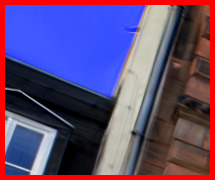}
\includegraphics[width=0.09478\textwidth]{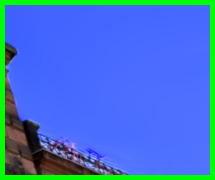}
\includegraphics[width=0.09478\textwidth]{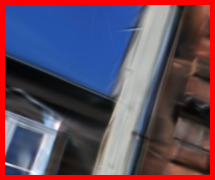}
\includegraphics[width=0.09478\textwidth]{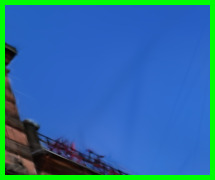}
\includegraphics[width=0.09478\textwidth]{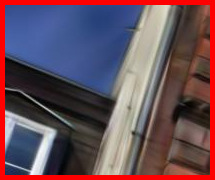}
\includegraphics[width=0.09478\textwidth]{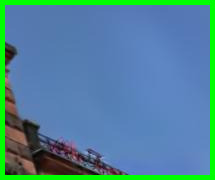}
\includegraphics[width=0.09478\textwidth]{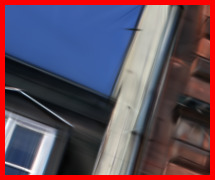}
\includegraphics[width=0.09478\textwidth]{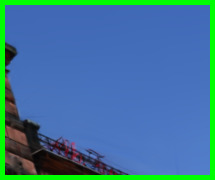}
\includegraphics[width=0.09478\textwidth]{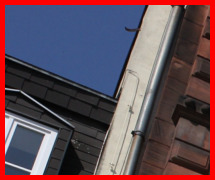}
\includegraphics[width=0.09478\textwidth]{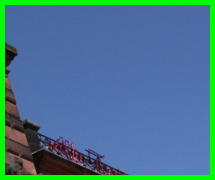}
\\

    \vspace{1em}
    \includegraphics[width=0.195\textwidth]{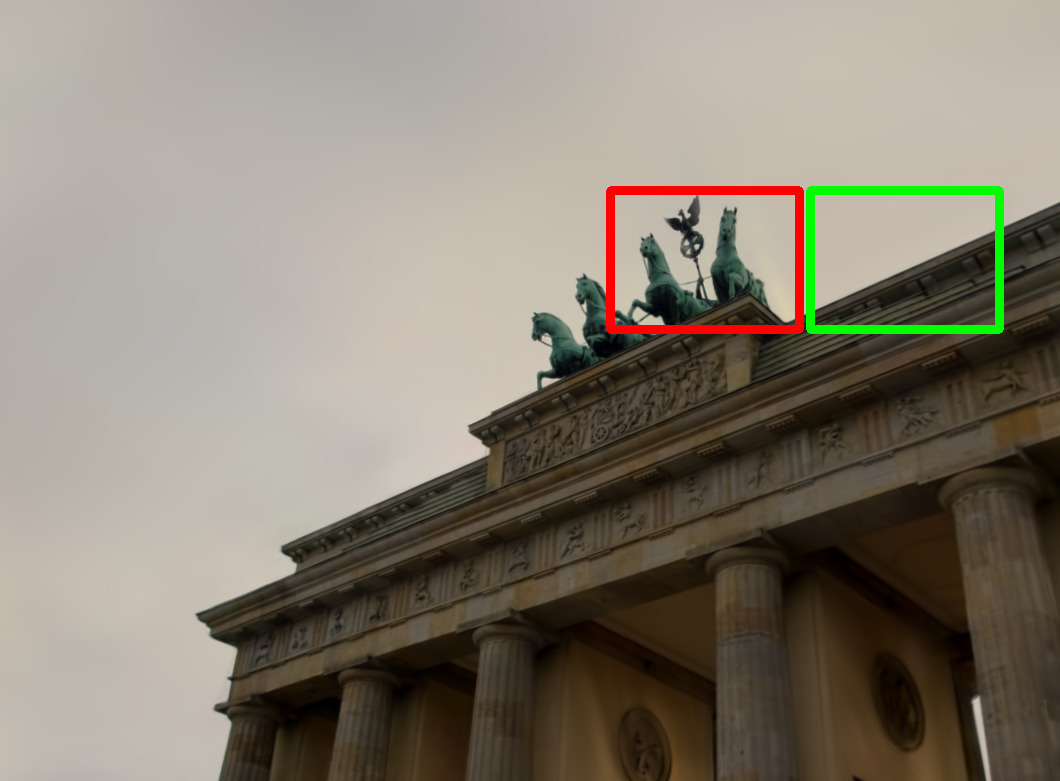}
\includegraphics[width=0.195\textwidth]{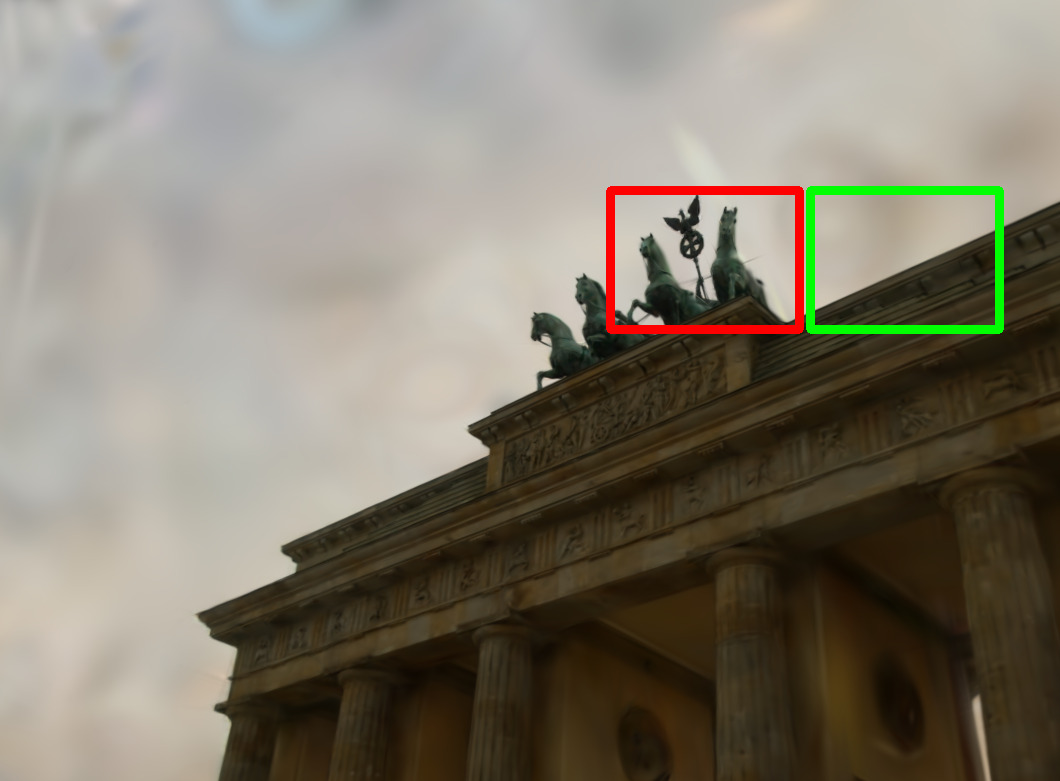}
\includegraphics[width=0.195\textwidth]{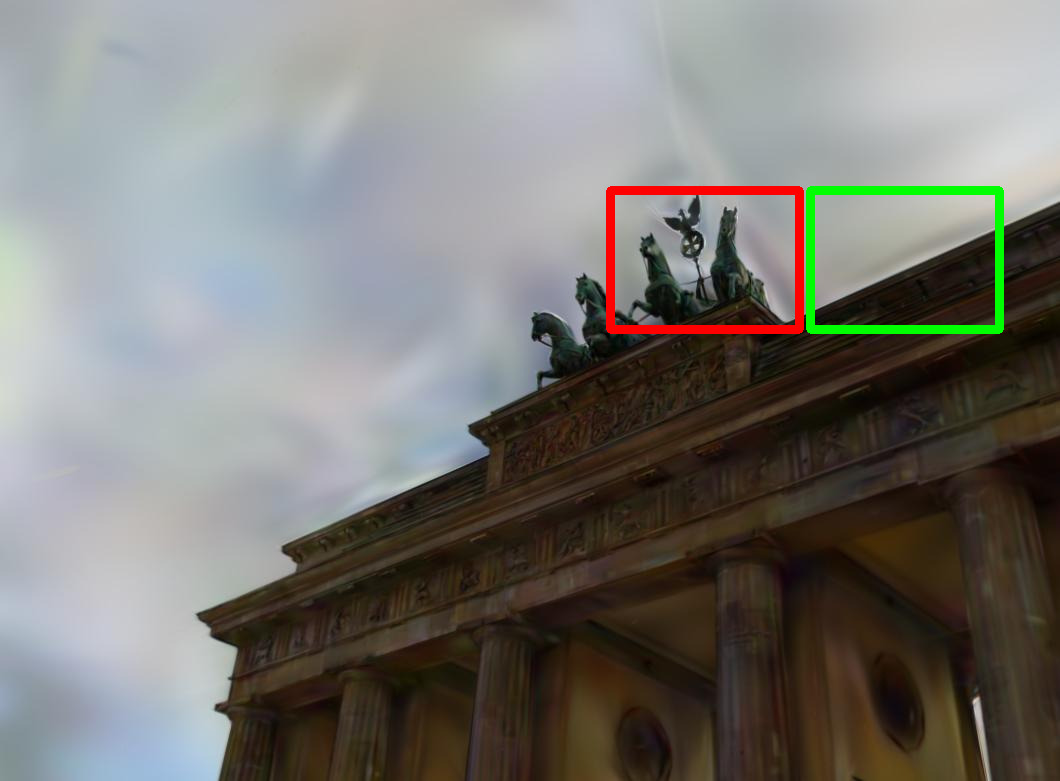}
\includegraphics[width=0.195\textwidth]{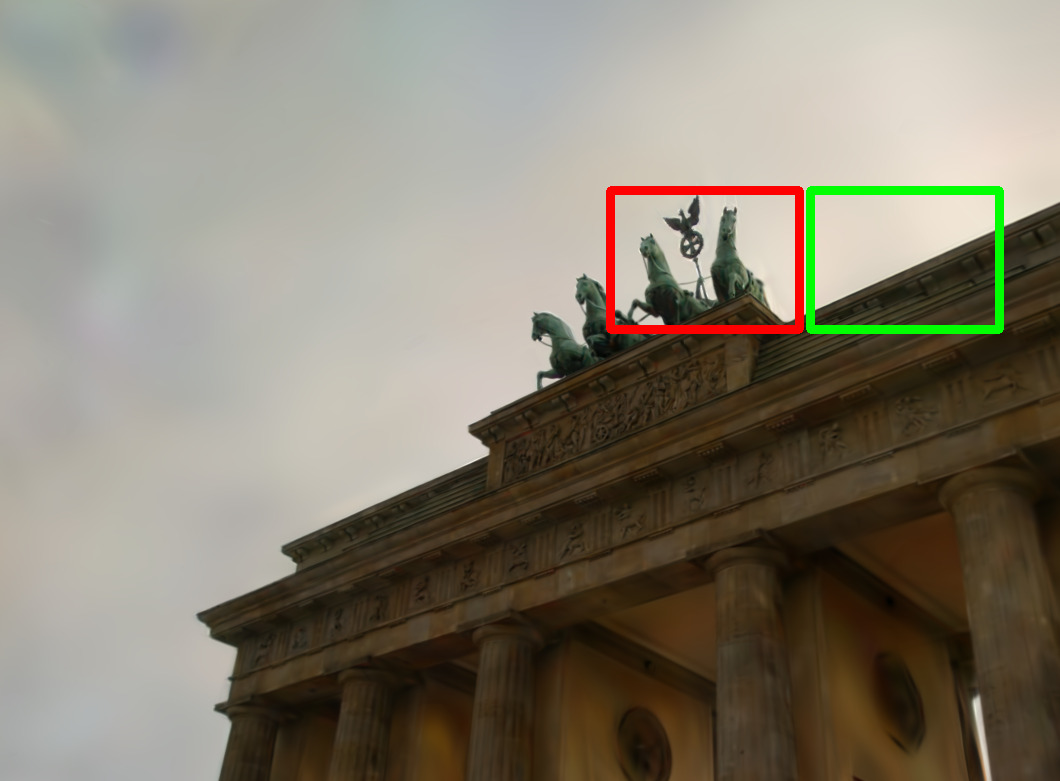}
\includegraphics[width=0.195\textwidth]{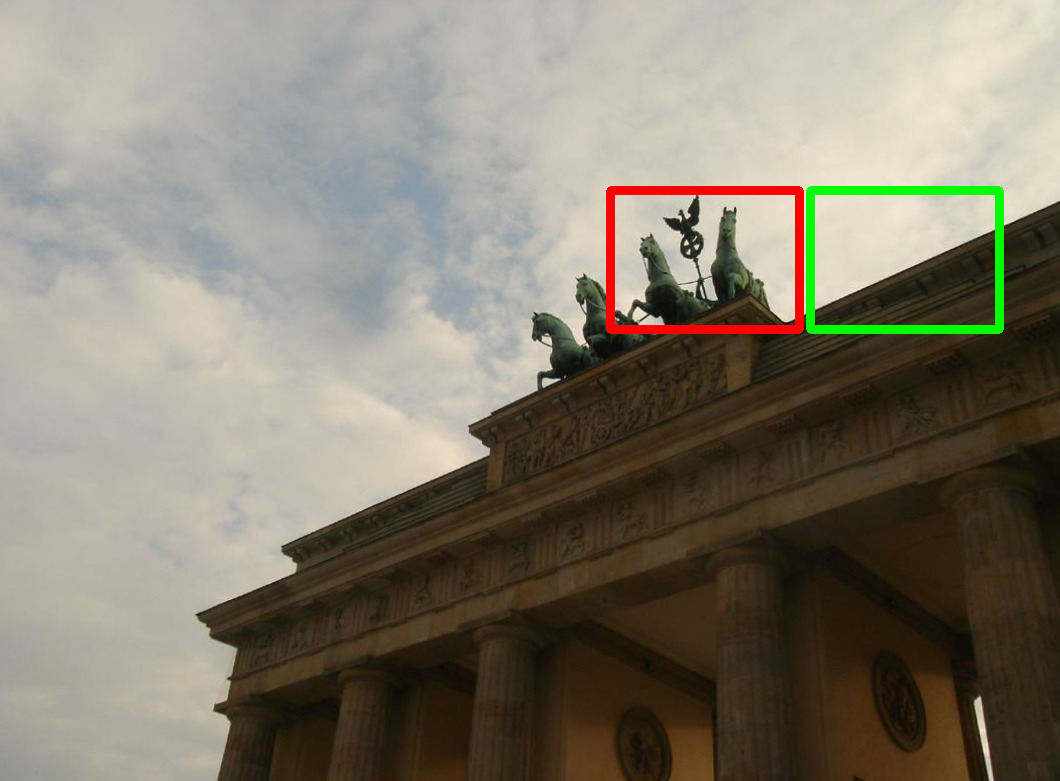}
\\
\includegraphics[width=0.09478\textwidth]{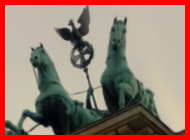}
\includegraphics[width=0.09478\textwidth]{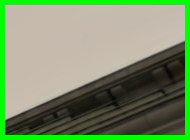}
\includegraphics[width=0.09478\textwidth]{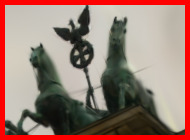}
\includegraphics[width=0.09478\textwidth]{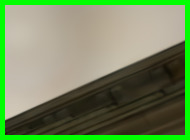}
\includegraphics[width=0.09478\textwidth]{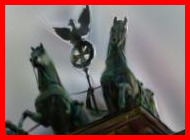}
\includegraphics[width=0.09478\textwidth]{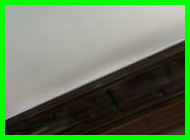}
\includegraphics[width=0.09478\textwidth]{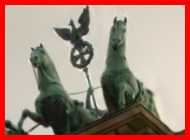}
\includegraphics[width=0.09478\textwidth]{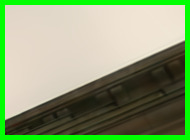}
\includegraphics[width=0.09478\textwidth]{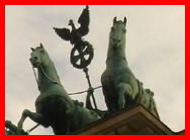}
\includegraphics[width=0.09478\textwidth]{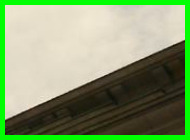}
\\[0.5em]
\includegraphics[width=0.195\textwidth]{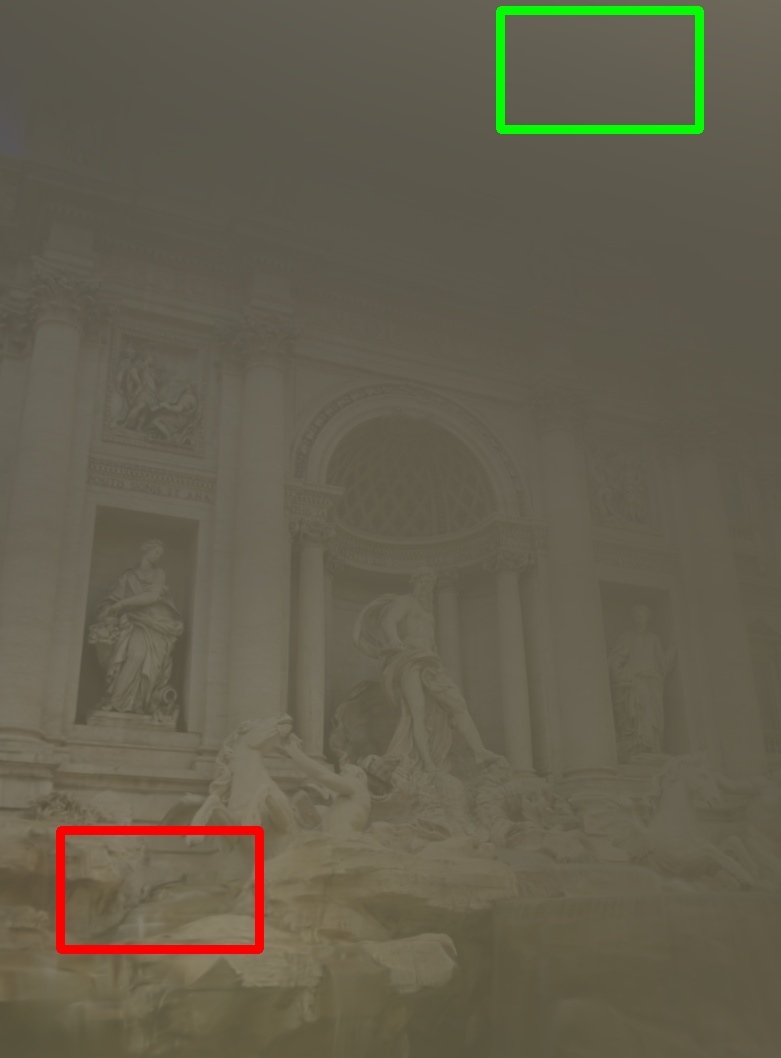}
\includegraphics[width=0.195\textwidth]{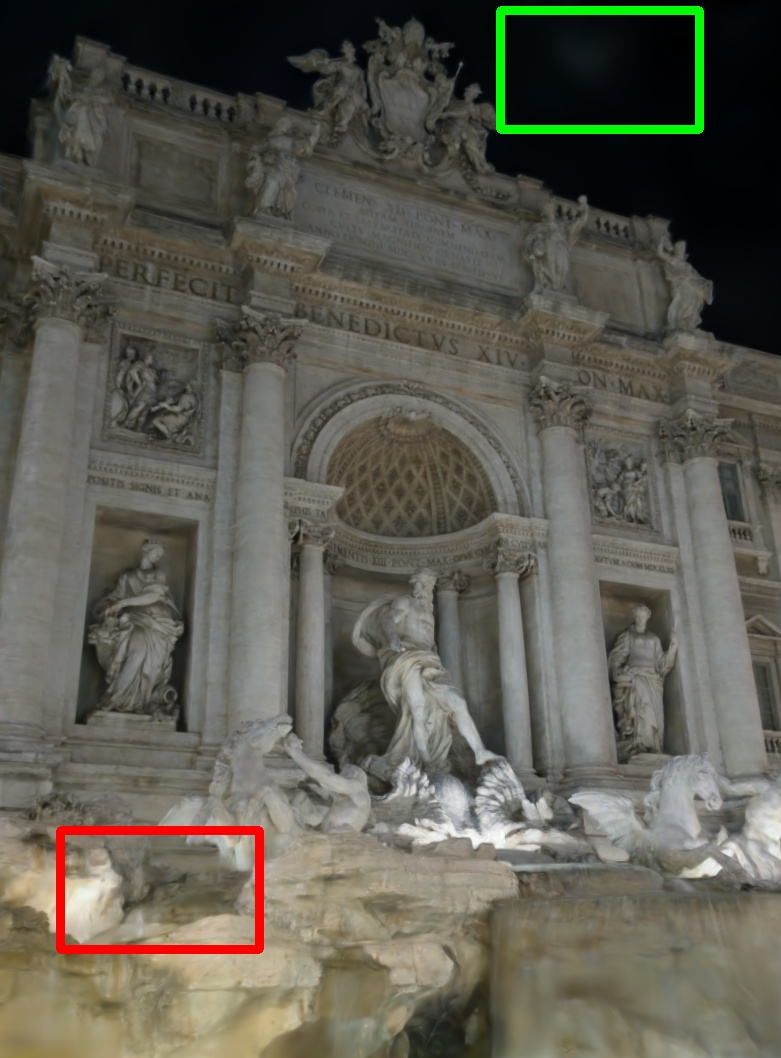}
\includegraphics[width=0.195\textwidth]{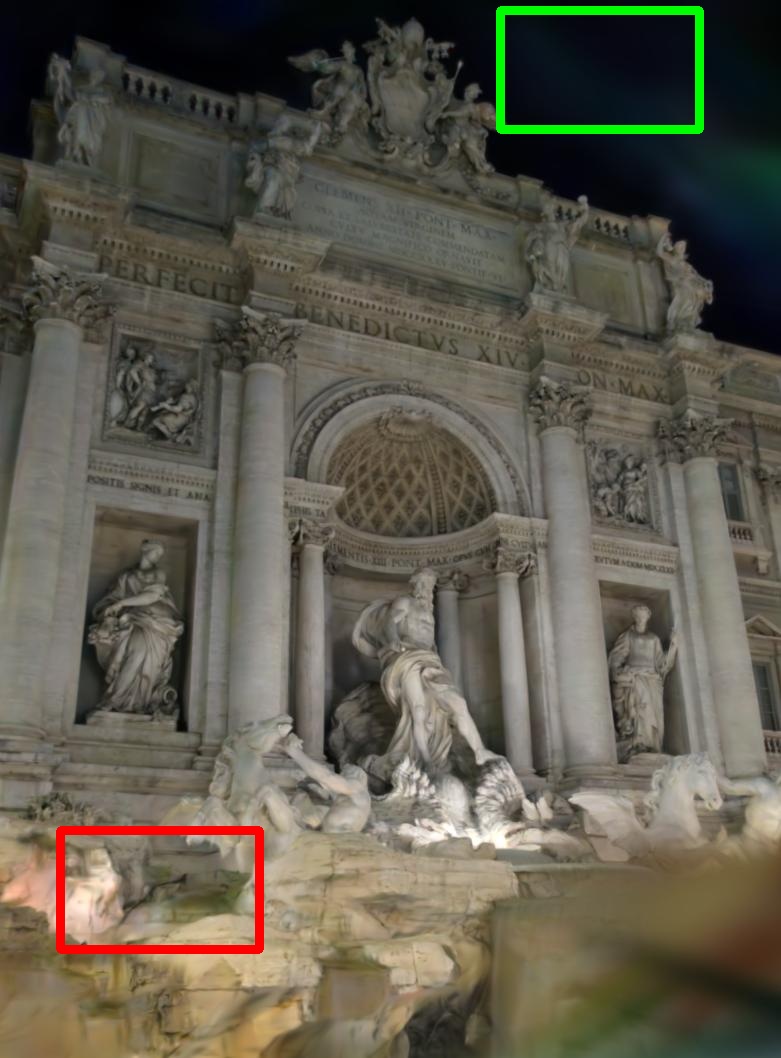}
\includegraphics[width=0.195\textwidth]{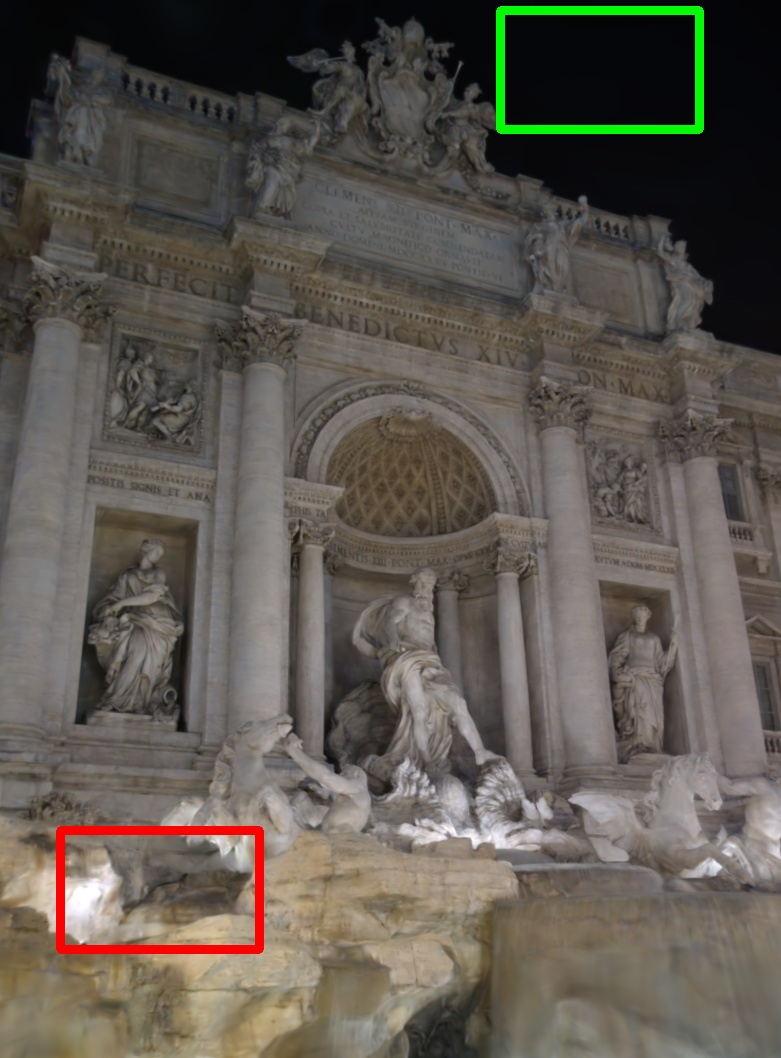}
\includegraphics[width=0.195\textwidth]{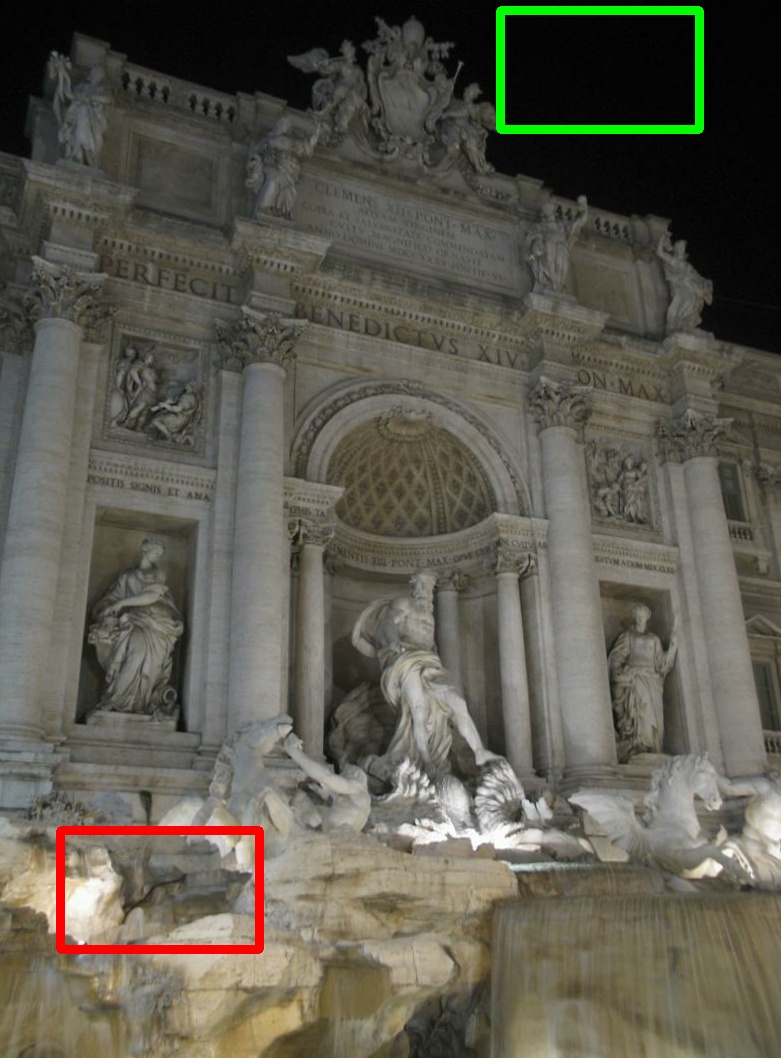}
\\
\includegraphics[width=0.09478\textwidth]{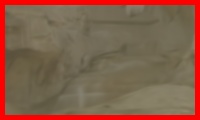}
\includegraphics[width=0.09478\textwidth]{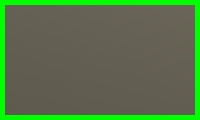}
\includegraphics[width=0.09478\textwidth]{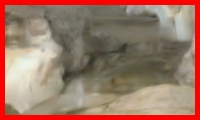}
\includegraphics[width=0.09478\textwidth]{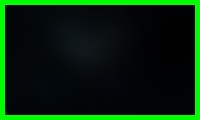}
\includegraphics[width=0.09478\textwidth]{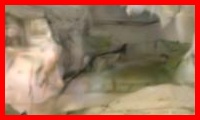}
\includegraphics[width=0.09478\textwidth]{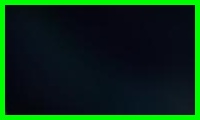}
\includegraphics[width=0.09478\textwidth]{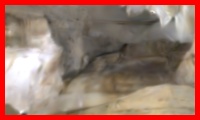}
\includegraphics[width=0.09478\textwidth]{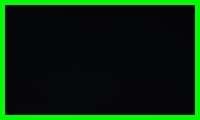}
\includegraphics[width=0.09478\textwidth]{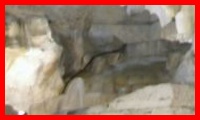}
\includegraphics[width=0.09478\textwidth]{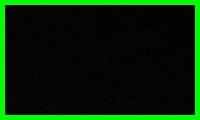}
\\

    \captionof{figure}{Qualitative comparison on \emph{lk2}, \emph{lwp} from NeRF-OSR dataset~\cite{rudnev2022nerfosr} and \emph{Brandenburg}, \emph{Trevi} from PhotoTourism dataset~\cite{2006phototourism}.}
    \label{fig:qual_osrpt}
\end{table*}

\subsection{Evaluation for \textbf{Unconstrained} 3DGS}

\paragraph{Baselines} 
{We evaluated our RobustSplat++ against multiple baselines, including the 3D Gaussian Splatting~\cite{kerbl20233d}, 3D Gaussian Splatting with exposure compensation~\cite{kerbl20233d} which we call} 3DGS-E, and recent robust methods including NexusSplats~\cite{ungermann2024robust}, WildGaussians~\cite{kulhanek2024wildgaussians} and DeSplat~\cite{wang2024desplat}.

{To ensure a fair comparison, 
all quantitative results were produced by optimizing the embeddings on the left half of the test images and evaluating exclusively on the right half. Visual comparisons are presented on the full test images.}

\paragraph{\textbf{\emph{NeRF-OSR}} Dataset}
We first evaluate our RobustSplat++ on the NeRF-OSR dataset. As shown in \Tref{tab:osrpt}, our approach achieves the best results on PSNR, SSIM, and LPIPS metrics across all three scenes, demonstrating the effectiveness of our method in handling illumination changes.

The qualitative comparisons are shown in \fref{fig:qual_osrpt}. Thanks to the proposed delayed Gaussian growth, our method successfully eliminates artifacts and achieves appearance closer to the ground truth  (e.g., the details of building in \emph{lk2} and the sky in \emph{lwp} scene) compared to state-of-the-art methods.

\paragraph{\textbf{\emph{PhotoTourism}} Dataset}
We also compare RobustSplat++ with state-of-the-art methods on the PhotoTourism dataset. 
The quantitative results in \Tref{tab:osrpt} show that our method achieves the most competitive performance compared to state-of-the-art approaches, despite performing slightly worse in PSNR on the \emph{Trevi Fountain} scene.
\fref{fig:qual_osrpt} presents the qualitative comparison, where the state-of-the-art approaches exhibit noticeable artifacts. In contrast, our method produces much cleaner results and achieves illumination colors consistent with the ground truth images.

\subsection{Ablation Study for \textbf{Unconstrained} 3DGS}
\label{sec:ablation_illumination}
{
To evaluate the effectiveness of each component of our method for modeling illumination variations, we built upon the RobustSplat for transient-free 3DGS and introduced different components to analyze the performance.
}

\paragraph{Effects of Appearance Modeling}
As shown in \Tref{table:ablation_illumination}, incorporating appearance modeling into RobustSplat leads to a significant performance improvement for handling illumination variations, resulting in the full RobustSplat++ model.

\paragraph{Effects of Mask Learning}
{The proposed mask learning is supervised by residuals and feature cosine similarity, which are affected by illumination changes. To assess the mask quality and its impact on appearance modeling, we disable mask learning and use ground-truth masks. As shown in \Tref{table:ablation_illumination}, the results of RobustSplat++ are comparable to the model with ground-truth masks, indicating that the learned mask is accurate enough and does not hinder illumination modeling.}

\paragraph{Effects of Delayed Gaussian Growth}
{Based on our observations, the Gaussian densification process has a strong connection to illumination-related artifacts. As shown in \Tref{table:ablation_illumination}, removing \delayedGS causes a clear drop in all metrics, indicating that \delayedGS helps suppress these artifacts during early densification. 
Since \delayedGS is} originally designed for transient handling, we also disable mask learning and use ground-truth masks to control variables. The results confirm that our analysis remains valid, further demonstrating the effectiveness of \delayedGS.

\begin{table}[t]
    \begin{center}
    \caption{{Ablation of each component in our appearance modeling on NeRF-OSR datasets~\cite{rudnev2022nerfosr}. ``RobustSplat'' is the baseline for transient-free 3DGS. We denote \emph{\DelayedGS} as ``DG'', and \emph{ground truth masks} as ``GM''. ``RobustSplat++ w/ GM'' indicates RobustSplat++ without masks of RobustSplat, while using ground truth mask provided by~\cite{rudnev2022nerfosr}.}}
    \label{table:ablation_illumination}
    \resizebox{0.48\textwidth}{!}{
    \begin{tabular}{l|cc|cc|cc}
    \toprule
    \multirow{2}{*}{Method} 
    & \multicolumn{2}{c|}{Site 1 (lk2))}  
    & \multicolumn{2}{c|}{Site 2 (st)} 
    & \multicolumn{2}{c}{Site 3 (lwp)} 
\\
    & PSNR & SSIM 
    & PSNR & SSIM 
    & PSNR & SSIM 
\\
    \midrule
    RobustSplat
    & 15.41 & 0.649 
    & 13.12 & 0.612
    & 11.06 & 0.563 
\\
    RobustSplat++ 
    & \textbf{19.13} & 0.707  
    & \textbf{16.54} & \textbf{0.667}
    & \textbf{14.88} & \textbf{0.648}  
\\
    RobustSplat++ w/ GM
    & \textbf{19.13} & \textbf{0.708}
    & 16.36 & \textbf{0.667}
    & 14.83 & 0.644 
\\
    RobustSplat++ w/o DG
    & 18.22 & 0.687 
    & 15.80 & 0.657
    & 14.54 & 0.638 
\\
    RobustSplat++ w/o DG \& w/ GM
    & 17.73 & 0.678 
    & 16.16 & 0.664
    & 14.47 & 0.636 
\\

    \bottomrule
    \end{tabular}}

    \end{center}
\end{table}

\section{Conclusion}
\label{sec:Conclusion}
In this work, we introduce RobustSplat{++}, a robust framework for {in-the-wild} 3D Gaussian Splatting, effectively mitigating artifacts caused by transient objects {and illumination variations}. 
Building on our analysis of the relation between Gaussian densification and artifacts caused by transient objects {and illumination variations}, our approach integrates a delayed Gaussian growth strategy to prioritize static scene optimization and a scale-cascaded mask bootstrapping method for reliable transient object suppression, while integrating appearance modeling to handle illumination variations. Through comprehensive experiments on multiple challenging datasets, RobustSplat{++} demonstrates superior robustness and rendering quality compared to existing methods.

\paragraph{Limitations} 
{Our current approach enables novel-view synthesis and 3D modeling from unconstrained image collections in in-the-wild scenes, but handling scene geometry changes under spatio-temporal variations remains highly challenging. In addition, our embedding-based appearance modeling requires test-time optimization on a partial region of unseen views, which limits the generalization and real-time performance of our approach. In future work, we aim to explore solutions for modeling spatio-temporal dynamics to achieve more robust reconstruction in uncontrolled environments.}

\bibliographystyle{IEEEtran}
\bibliography{ref}

\vfill

\end{document}